\documentclass{article} 
\usepackage{iclr2021_conference,times}

\usepackage{amsmath,amsfonts,bm}

\def\eqref#1{equation~\ref{#1}}

\def\1{\bm{1}}

\DeclareMathAlphabet{\mathsfit}{\encodingdefault}{\sfdefault}{m}{sl}
\SetMathAlphabet{\mathsfit}{bold}{\encodingdefault}{\sfdefault}{bx}{n}

\DeclareMathOperator*{\argmax}{arg\,max}

\usepackage{hyperref}
\usepackage{url}

\usepackage[utf8]{inputenc} 
\usepackage[T1]{fontenc}    
\usepackage{booktabs}       
\usepackage{amsfonts}       
\usepackage{nicefrac}       
\usepackage{microtype}      
\usepackage{wrapfig}
\usepackage{amsmath}
\usepackage{amsthm}
\usepackage{bbm}
\usepackage{amssymb}
\usepackage{mathtools}
\usepackage{multirow}
\usepackage{amsmath,amsfonts,amssymb, amsthm}
\usepackage{mathtools}
\usepackage{fixmath}
\usepackage{caption}
\usepackage{multirow, makecell}
\usepackage{float}
\usepackage{enumitem}
\usepackage{boxedminipage}
\usepackage{dashbox}
\usepackage{arydshln}
\usepackage{tikz}
\usepackage{pgfplots}
\usepackage{tabularx}
\newcommand{\rtable}[1]{\renewcommand{\arraystretch}{#1}}
\newcommand\Tstrut{\rule{0pt}{2.6ex}}         
\newcommand\Bstrut{\rule[-1.2ex]{0pt}{0pt}}   
\newcommand\TBstrut{\Tstrut\Bstrut}           

\usepackage{thmtools}
\usepackage{thm-restate}
\usepackage{graphicx,subfig}
\newcommand{\LineSep}[1][-0.45]{%
  \par\vspace*{\dimexpr-\baselineskip}%
}
\usepackage{cleveref}


\title{Improving Local Effectiveness for \\Global robust training}

\author{Jingyue Lu \\
  University of Oxford\\
  \texttt{jingyue.lu@spc.ox.ac.uk} \\
\And
 M. Pawan Kumar \\
       University of Oxford\\
  \texttt{pawan@robots.ox.ac.uk} \\
}

\iclrfinalcopy 
\begin{document}

\maketitle
\vspace{-10pt}
\begin{abstract}
\vspace{-10pt}
Many successful robust training methods rely on strong adversaries, which can be prohibitively expensive to generate when the input dimension is high and the model structure is complicated. We adopt a new perspective on robustness and propose a novel training algorithm that allows a more effective use of adversaries. Our method improves the model robustness at each local ball centered around an adversary and then, by combining these local balls through a global term, achieves overall robustness. Focusing on local balls improves the efficiency of robust training. We demonstrate the performance of our method on MNIST, CIFAR-10, CIFAR-100. 
\vspace{-10pt}
\end{abstract}

\section{Introduction}
\vspace{-10pt}
Despite their popularity, naturally trained deep neural networks (DNNs) are fragile. By adding to each data a perturbation that is carefully designed but imperceptible to humans, DNNs previously reaching almost 100$\%$ accuracy performance could hardly make a correct prediction any more \citep{Szegedy2013}. To tackle these issues, training DNNs that are robust to small perturbations has become an active area of research in machine learning.

Various algorithms have been proposed \citep{distill, advlogit, trades, LLR, Curvature, PGD,MMA}. Among them, many methods require using strong adversarial attacks, generally computed through several steps of projected gradient descent. Such attacks can quickly become prohibitive when model complexity and input dimensions increase, 
thereby limiting their applicability. Since the cost of finding strong adversaries is mainly due to the high number of gradient steps performed, one potential approach to alleviate the problem is to use cheap but weak adversaries. Weak adversaries are obtained using fewer gradient steps, and in the extreme case with a single gradient step. Based on this idea, \citet{wong2020fast} argue that by using random initialization and a larger step-size, adversarial training with weak adversaries found via one gradient step is sufficient to achieve a satisfactory level of robustness. We term this method as one-step ADV from now on. While one-step ADV does indeed exhibit robustness, there is still a noticeable gap when compared with its multi-step counterpart.

In this paper, we further bridge the gap by proposing a new robust training algorithm: Adversarial Training via LocAl Stability (ATLAS). Local stability, in our context, implies stability of prediction and is the same as local robustness. We show that ATLAS makes a more effective use of weak adversaries by favourably comparing it against one-step ADV on three datasets: MNIST, CIFAR-10 and CIFAR-100. When strong adversaries are used, ATLAS matches with the current state of the art on MNIST and outperforms them on CIFAR-10 and CIFAR-100.

\vspace{-10pt}
\section{Related Works}
\vspace{-10pt}
Robust training aims to learn a network such that it is able to give the same correct output even when the input is slightly perturbed. Various types of robust training methods have been proposed. In this context, we focus on adversary based robust training, as they generally perform better in terms of robust accuracy. Under this category, an early fundamental work is the Fast Gradient Sign Method (FGSM) by \citet{Goodfellow2015}. Adversarial Training (ADV) \citep{PGD} is a multi-step variant of FGSM. Rather than using one step FGSM, ADV employs multi-step projected gradient descent (PGD) \citep{BIM} with smaller step-sizes to generate perturbed inputs. These modifications have allowed ADV to become one of the most effective robust training methods so far \citep{athalye2018obfuscated}. Another frequently used robust training method is TRADES \citep{trades}. TRADES encourages model robustness by adding to the natural loss a regularizer involving adversaries to push away the decision boundary. Recently, \citet{LLR} suggest that a robust model can be learned through promoting linearity in the vicinity of the training examples. They designed a local linearity regularizer (LLR), in which adversaries are used to maximize the penalty for non-linearity. Applying LLR also allows efficient robust training. We note that the underlying idea of LLR is complementary to ATLAS.

One major drawback of adversary based methods \citep{PGD,trades,LLR,wang2020misclass, ding2019mma} is that most of them rely on strong adversaries, computed via PGD. When the input dimension is high and the model structure is complicated, finding adversaries can be too expensive for these methods to work effectively. Several works have researched possible ways to speed up the process. Algorithmically, \citet{zhang2019yopo} cut down the total number of full forward and backward passes for generating multi-step PGD adversaries while \citet{zhang2020attacks} introduce a parameter to allow early stop PGD. \citet{freeadv} reduce adversary overhead by combining weight update and input perturbation update within a single back propagation and use a single step FGSM adversary. \citet{wong2020fast} argue that the main reason one-step FGSM, generally regarded as not robust, is effective in \citep{freeadv} is due to the non-zero initialization used. As a result, \citet{wong2020fast} proposed that with random initialization and a larger step size, weak adversaries generated by FGSM could lead to models with a high level of robust accuracy. In this study, we adopt a similar viewpoint of accelerating robust training through the use of weak adversaries. 
\vspace{-10pt}
\section{ATLAS}
\vspace{-10pt}
Before we present our method---Adversarial Training via LocAl Stability (ATLAS)---we introduce our problem set-up. We consider classification tasks. Let $\boldsymbol{x}$ be an image
from the data distribution $\mathcal{X}\subset \mathbb{R}^{N}$ and $y_{\boldsymbol{x}}$ be its correct label from classes $\mathcal{C}=\{1,\dots,C\}$. We denote a neural network, parameterized by $\boldsymbol{\theta}$, as $f(\boldsymbol{x};\boldsymbol{\theta}) :\, \mathcal{X}\rightarrow \mathbb{R}^C.$ The function $f$ outputs logits, using which the prediction probabilities $p(\boldsymbol{x};\boldsymbol{\theta})$ are computed as $p_i(\boldsymbol{x}; \boldsymbol{\theta}) = \frac{\exp (f_i(\boldsymbol{x};\boldsymbol{\theta}))}{ \sum_j\exp(f_j(\boldsymbol{x};\boldsymbol{\theta}))}$ element-wise. In addition, we use a ball $\mathcal{B}_{\epsilon}(\boldsymbol{x}) = \{\boldsymbol{x}' \,|\, \Vert \boldsymbol{x}'-\boldsymbol{x} \Vert_{p} \leq \epsilon\}$ to represent the allowed perturbation. If $f$ is robust on $\mathcal{B}_{\epsilon}$, that is, it predicts the same class for all $\boldsymbol{x}' \in {\cal B}_\epsilon$, we call $\mathcal{B}_{\epsilon}$ a robust ball.

To facilitate the discussion, we further introduce the following notation. We assume that the neural network $f$ contains ReLU activation for the sake of clarity. This implies that the neural network is piecewise linear. As a consequence, at each $\boldsymbol{x}\in\mathcal{X}$, it is easy to find a weight matrix $W^{\boldsymbol{x}}\in\mathbb{R}^{\mathcal{C}\times N}$ and a constant vector $\boldsymbol{b}^{\boldsymbol{x}}\in\mathbb{R}^{\mathcal{C}}$ such that $f(\boldsymbol{x}; \boldsymbol{\theta}) = W^{\boldsymbol{x}}\boldsymbol{x} +\boldsymbol{b}^{\boldsymbol{x}} $. To simplify the notation, we define
\begin{equation}\label{w-def}
   \breve{W}^{\boldsymbol{x}} = [\, W^{\boldsymbol{x}}\, |\,\boldsymbol{b}^{\boldsymbol{x}} \,] \in \mathbb{R}^{\mathcal{C} \times (N+1)}, \qquad \breve{\boldsymbol{x}}^{T} = [\, \boldsymbol{x}^{T} \,|\,1\,] \in \mathbb{R}^{1\times (N+1)}.
\end{equation}
As a result, at each point $\boldsymbol{x}$, we have $f(\boldsymbol{x};\boldsymbol{\theta}) = \breve{W}^{\boldsymbol{x}}\breve{\boldsymbol{x}}$.
\vspace{-5pt}
\subsection{binary classification problem}
\begin{figure}[h]
 \vspace{-10pt}
  \begin{minipage}[c]{0.55\textwidth}
     \includegraphics[width=0.95\textwidth]{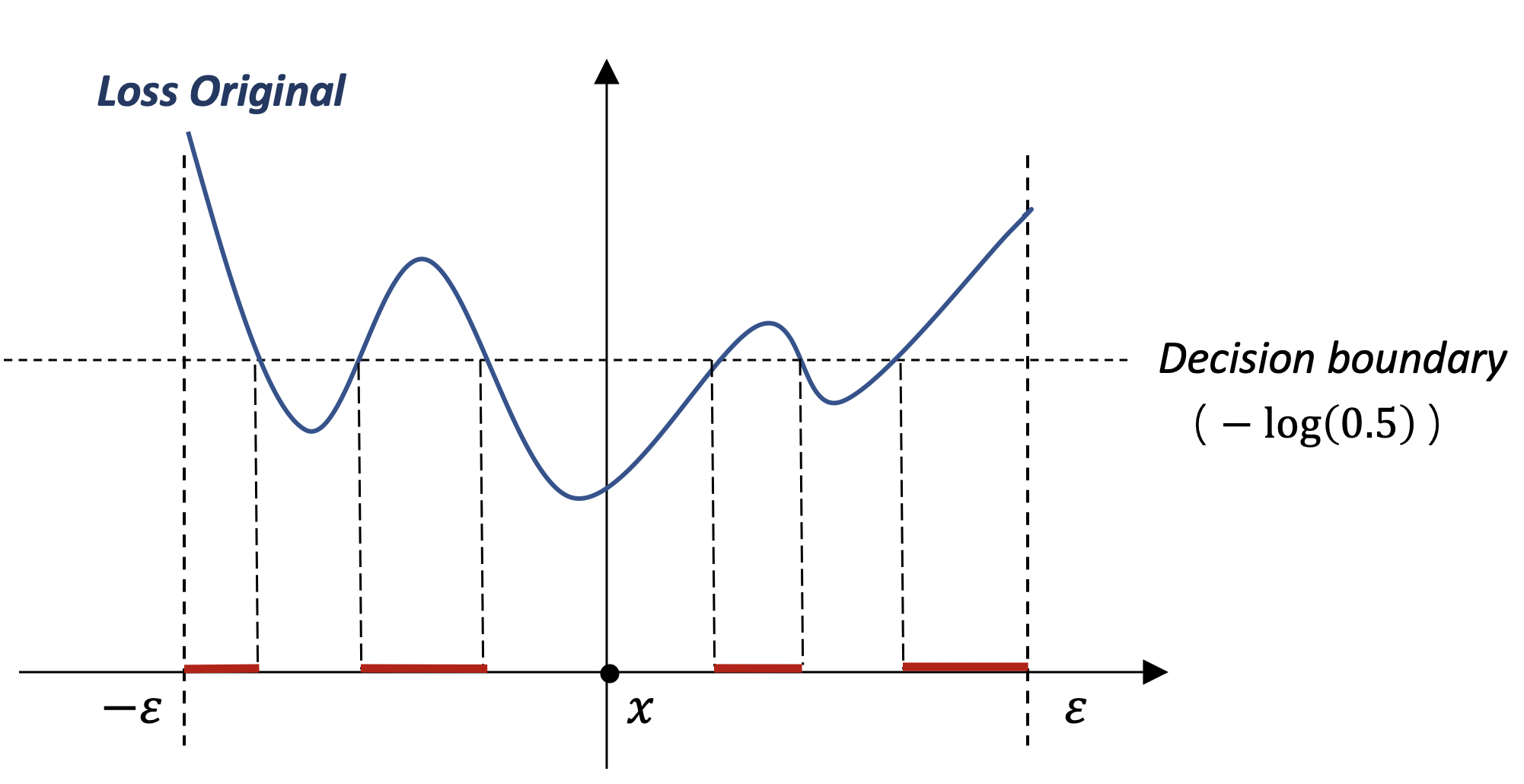}
  \end{minipage}\hfill
  \begin{minipage}[c]{0.4\textwidth}
    \caption{\label{fig:loss-original} Original loss curve on $\mathcal{B}_{\epsilon}(\boldsymbol{x})$ by fixing the value of $\boldsymbol{x}$ on all dimensions apart from one. Points with the loss above the decision boundary $-\log{0.5}$ are adversarial points and marked as red. }
  \end{minipage}
   \vspace{-10pt}
\end{figure}

\vspace{-5pt}
We illustrate the key idea of ATLAS through a binary classification problem. Let there be two classes $\mathcal{C}=\{c_1,c_2\}$. For an arbitrary point $\boldsymbol{x}\in \mathcal{X}$, the correct class label is $y_{\boldsymbol{x}} = c_1$. Since we are interested in a network's performance in classifying labels, we compute logits difference as 
\begin{equation*}
f_1(\boldsymbol{x};\boldsymbol{\theta}) -f_2(\boldsymbol{x};\boldsymbol{\theta}) =  \boldsymbol{d}\breve{W}^{\boldsymbol{x}}\breve{\boldsymbol{x}} \in \mathbb{R},
\end{equation*}
where $\boldsymbol{d} = [1, -1]$ is a row vector. For a binary classification problem with cross-entropy loss, the loss decreases monotonically with the increase of the value $\boldsymbol{d}\breve{W}^{\boldsymbol{x}}
\breve{\boldsymbol{x}}$. To achieve the desired robustness, we require $\boldsymbol{d}\breve{W}^{\boldsymbol{x}'}\breve{\boldsymbol{x}}'>0$ for all $\boldsymbol{x}' \in\mathcal{B}_{\epsilon}(\boldsymbol{x})$. When cross entropy loss $\ell^{\text{CE}}$ is considered, we need $\ell^{\text{CE}}(\boldsymbol{x}')<-\log(0.5)$ for all $\boldsymbol{x}' \in\mathcal{B}_{\epsilon}(\boldsymbol{x})$. In Figure \ref{fig:loss-original}, we show a potential loss curve on $\mathcal{B}_{\epsilon}(\boldsymbol{x})$ by fixing the value of $\boldsymbol{x}$ on all dimensions apart from one. Our loss framework consists of a local component and a global component. Both components update the parameters of the network model. 
 
\subsection{Local Robustness}
\vspace{-5pt}
For the local component, at each given local point $\boldsymbol{x}'\in \mathcal{B}_{\epsilon}(\boldsymbol{x})$, the goal is to attain a robust ball $\mathcal{B}_{\gamma'}(\boldsymbol{x}')$ with large radius. We maximize the use of adversaries by treating them as local center points $\boldsymbol{x}'$. In other words, given an adversary $\boldsymbol{x}'\in\mathcal{B}_{\epsilon}(\boldsymbol{x})$, we need to satisfy two requirements: the model predicts the required label $y_{\boldsymbol{x}}$ at $\boldsymbol{x}'$ and the radius $\gamma'$ of $\mathcal{B}_{\gamma'}(\boldsymbol{x}')$ should be enlarged. When compared with ADV, the additional second requirement allows us to achieve improved robustness performance even when weak adversaries are used.   

\begin{figure}[h]
\vspace{-10pt}
  \centering
  \begin{minipage}[t]{0.48\textwidth}
    \includegraphics[width=0.95\textwidth]{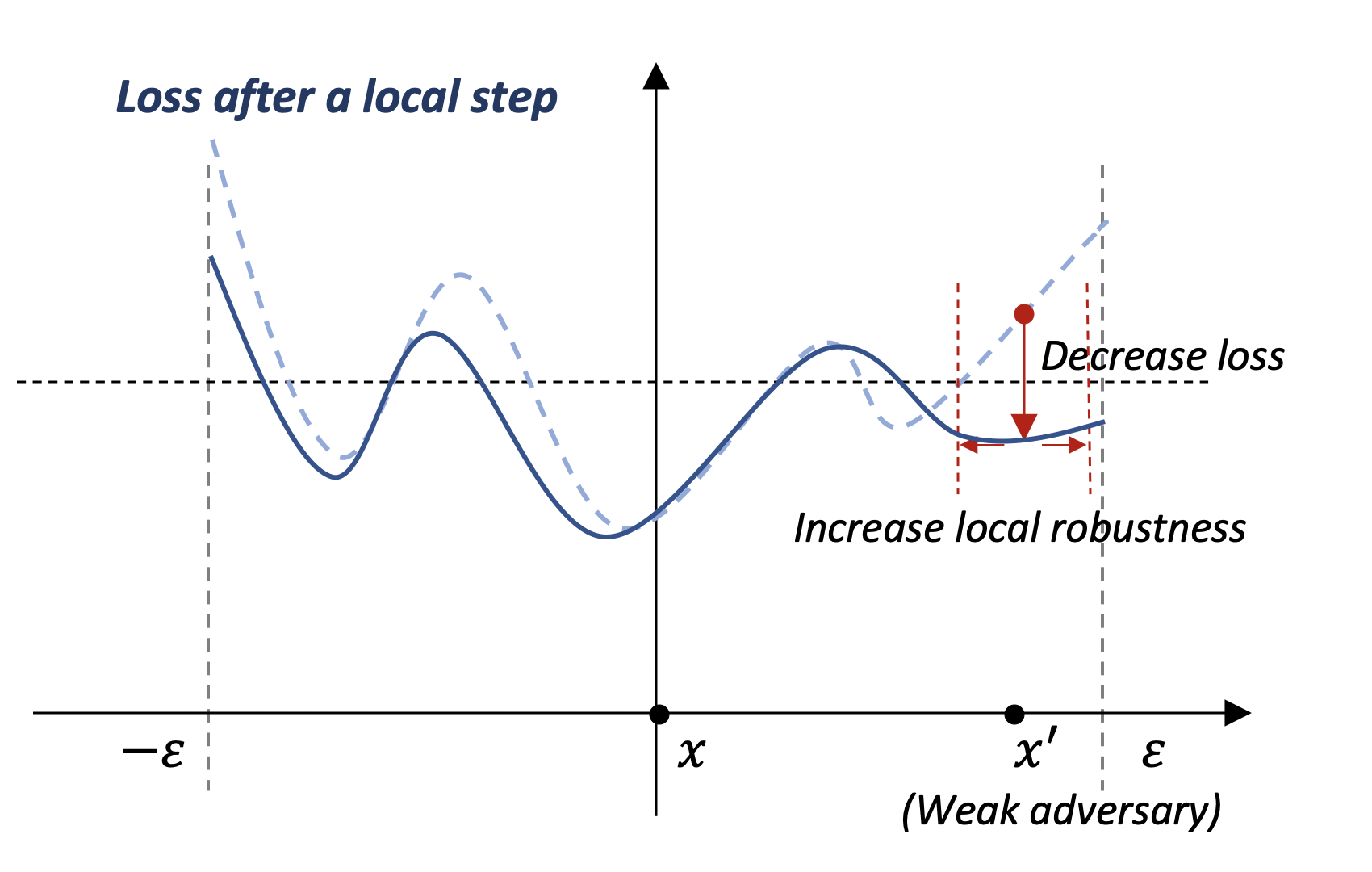}
    \caption{\label{fig:loss-local} Loss curve after a local step. The dashed line is the original loss curve. Provided with a weak adversary $\boldsymbol{x}'$, the local step aims to decrease the loss at $\boldsymbol{x}'$ while increasing local robustness so the local robust ball centered at $\boldsymbol{x}'$ has a large radius.}
  \end{minipage}
   \hfill
  \begin{minipage}[t]{0.48\textwidth}
    \includegraphics[width=0.95\textwidth]{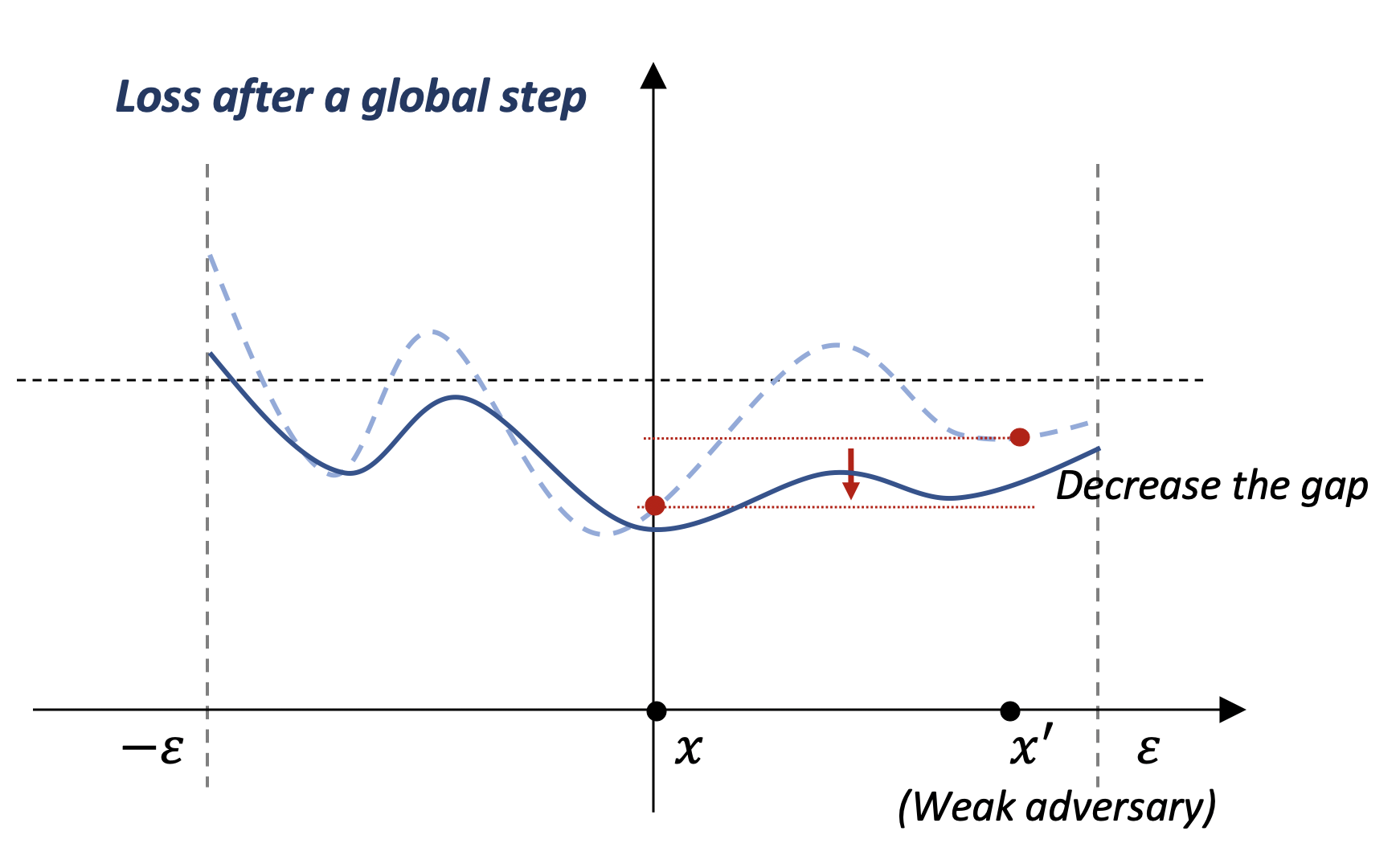}
    \caption{\label{fig:loss-global} Loss curve after a global step. The dashed line is the loss curve after a local step. To further smooth the loss curve and to remove adversaries between $\boldsymbol{x}$ and $\boldsymbol{x}'$, we penalize the difference $p(\boldsymbol{x};\boldsymbol{\theta})-p(\boldsymbol{x}';\boldsymbol{\theta})$. As a result, the gap between $\ell(\boldsymbol{x})$ and $\ell(\boldsymbol{x}')$ is decreased. }
  \end{minipage}
  \vspace{-10pt}
\end{figure}

The first requirement can be easily met by applying a standard loss term at $\boldsymbol{x}'$ with the label $y_{\boldsymbol{x}}$. For the second requirement, to push away the decision boundary, various methods including Jacobian regularizers \citep{jac2018jakubovitz, ross2018improving}, MMA \citep{ding2019mma} and MMR \citep{MaxMargin} have been introduced. An expected loss curve after introducing the local component is shown in Figure \ref{fig:loss-local}. Given the goal of efficient robust training and the fact that an efficient estimation algorithm for Jacobian has been developed in \citet{Jac}, we use the approximated Jacobian term $\Vert J(\boldsymbol{x}')\Vert_{F}^{approx}$ to meet the local component requirements. 
\vspace{-5pt}
\subsection{Global Regularization}
\vspace{-5pt}
In terms of the global component, we combine local balls together in a controlled way to further improve robustness. The goal is that, with the global regularization step, points that are misclassified even after the local step can be correctly predicted, as illustrated in Figure \ref{fig:loss-global}. To achieve this goal, our analysis (provided in the appendix) shows it is sufficient to minimise the distance between $\boldsymbol{d}\breve{W}^{\boldsymbol{x}}\breve{\boldsymbol{x}}$ and $\boldsymbol{d}\breve{W}^{\boldsymbol{x}'}\breve{\boldsymbol{x}}'$. We thus equivalently penalize the KL distance of $p(\boldsymbol{x};\boldsymbol{\theta})-p(\boldsymbol{x}';\boldsymbol{\theta})$ for the global part. 

When the loss surface is considered, the global component loss encourages the loss surface to be smoothened and flattend on $\mathcal{B}_{\epsilon}(\boldsymbol{x})$. As suggested in \citep{trades, cisse2017parseval, ross2018improving, moosavi2017analysis}, these properties of the loss surface are desired and are possibly indispensable for models to be robust. 
\vspace{-5pt}
\subsection{Final Loss}
\vspace{-5pt}
Integrating the above analyses, we propose the following as the final loss for ATLAS. Given $\alpha>0$ and $\beta >0$, the loss is formulated as
\begin{equation}
    \ell_{\text{ATLAS}} = \frac{1}{|\mathcal{X}|}\sum_{\boldsymbol{x}\in\mathcal{X}} \underbrace{\ell^{\text{CE}}(\boldsymbol{x}') + \alpha \cdot \Vert J(\boldsymbol{x}')\Vert_{F}^{approx}}_{\text{local}} + \underbrace{\beta \cdot \text{KL}(p(\boldsymbol{x}';\boldsymbol{\theta})\|p(\boldsymbol{x};\boldsymbol{\theta}))}_{\text{global}}, \quad \boldsymbol{x}'\in \mathcal{B}_{\epsilon}(\boldsymbol{x}).
\end{equation}
More details and analytical reasoning for ATLAS are provided in the appendix.
\vspace{-10pt}
\section{Experiments}
\vspace{-10pt}
We evaluate the performance of ATLAS on three datasets: MNIST, CIFAR-10 and CIFAR-100. We consider the cases of training robustness models with weak adversaries generated by FGSM and with strong adversaries by multi-step PGD. To be consistent with the general experimental setting, we adopt $l_\infty$ norm and use the perturbation value $\epsilon=0.3$ for MNIST and $\epsilon=8/255$ for CIFAR-10 and CIFAR-100. \textbf{Methods } We compare against ADV and TRADES. We term methods that use weak FGSM adversaries as \textbf{one-step methods}. One-step ADV is an implementation of \citep{wong2020fast}. \textbf{Multi-step methods} are trained on strong adversaries generated by PGD: 20 steps for the MNIST and 10 steps for CIFAR-10 and CIFAR-100. We also include the Jacobian penalty loss \citep{jac2018jakubovitz, Jac, ross2018improving}. We refer to it as zero-step JAC to emphasize the fact it does not require adversaries. \textbf{Robustness Evaluation } For evaluations, black box attack generates adversaries on surrogate models with 1000 PGD steps. In terms of white box attacks, apart from PGD-20, we further introduce three strong white attacks: PGD-1000, Untargeted (Un-T) attack \citep{Unt} and Multi-targeted (Multi-T) attack \citep{MultT}. More training and evaluation related details can be found in the appendix. 

\begin{table*}[t!]
\centering
	\rtable{1.2}
			\caption{Robustness performance for MNIST on a small CNN and for CIFAR-10 and CIFAR-100 on Wide-Resnet 28-8. The higher the better. \label{table:mainTable}}
			\resizebox{\linewidth}{!}{
			\begin{tabular}{clccccccccc}
				\toprule
				{} & {} & {}  & {}& \multicolumn{4}{c}{White box attack}  & \multicolumn{1}{c}{Time}\\
				\cmidrule(lr){5-8}\cmidrule(lr){9-9}
				& Model & \vtop{\hbox{\strut Clean}\hbox{\strut accuracy }} & \vtop{\hbox{\strut Black box}\hbox{\strut attack}}& \vtop{\hbox{\strut PGD-20}\hbox{\strut attack}} & \vtop{\hbox{\strut PGD-1000}\hbox{\strut attack}} & \vtop{\hbox{\strut Un-T }\hbox{\strut attack}} &\vtop{\hbox{\strut Multi-T}\hbox{\strut attack}}   & per batch \\
				\midrule
				\parbox[t]{2mm}{\multirow{8}{*}{\rotatebox[origin=c]{90}{MNIST}}}
				& zero-step JAC & 98.34$\%$ & 89.84$\%$ & 58.09$\%$ & 2.48$\%$ & 28.83$\%$ & 28.72$\%$  & 0.012s \\
				\cline{2-9}
				&one-step ADV & $\boldsymbol{99.49}\%$ & 96.26 $\%$  & 96.29$\%$  & 83.25$\%$ & 91.02$\%$& 90.62$\%$& 0.009s  \\
                &one-step TRADES & 99.40$\%$ & 96.21$\%$ & 96.28$\%$ & 83.05$\%$  & 90.66$\%$& 90.02$\%$ & 0.016s \\
                &one-step ATLAS   & 99.47$\%$ & $\boldsymbol{96.36}\%$ & $\boldsymbol{96.84}\%$& $\boldsymbol{89.56}\%$ &$\boldsymbol{92.44}\%$  & $\boldsymbol{92.04}\%$  & 0.018s\\
				\cline{2-9}
				&twenty-step ADV & 99.48$\%$& 96.40$\%$  & 97.16$\%$ & $\boldsymbol{92.32}\%$ & 93.01$\%$& $\boldsymbol{92.87}\%$& 0.077s \\
                &twenty-step TRADES & $\boldsymbol{99.49}\%$  &  96.21$\%$ & 97.01$\%$ & 90.51$\%$ & 92.49$\%$& 92.21$\%$ & 0.097s\\
                &twenty-step ATLAS & 99.44$\%$ &$\boldsymbol{96.45}\%$  & $\boldsymbol{97.19}\%$ & 92.20$\%$ & $\boldsymbol{93.03}\%$ &92.74$\%$ & 0.086s  \\
				\hline \hline \TBstrut\TBstrut
				
				\parbox[t]{2mm}{\multirow{8}{*}{\rotatebox[origin=c]{90}{CIFAR-10}}}
				&  zero-step JAC & 63.24$\%$& 60.45$\%$ & 34.40$\%$ & 34.34$\%$ & 31.13$\%$ & 30.32$\%$  & 0.47s \\
				\cline{2-9}
				& one-step ADV & 86.24$\%$ & 82.53$\%$ &45.73$\%$  & 44.98$\%$  & 45.14$\%$& 42.97$\%$  & 0.25s \\
                & one-step TRADES & $\boldsymbol{86.84}\%$ &$\boldsymbol{82.63}\%$ & 40.03$\%$ & 39.41$\%$ & 39.31$\%$  & 38.14$\%$& 0.55s \\
                & one-step ATLAS  & 84.52$\%$ &81.06$\%$ & $\boldsymbol{49.01}\%$& $\boldsymbol{48.59}\%$ & $\boldsymbol{48.54}\%$ & $\boldsymbol{46.55}\%$&  0.73s  \\
				\cline{2-9}
				 &ten-step ADV & $\boldsymbol{84.77}\%$ & 82.53$\%$ & 50.60$\%$ & 50.22$\%$ & 49.74$\%$& 47.96$\%$ & 1.38s\\
                 &ten-step TRADES & 84.01$\%$ & $\boldsymbol{82.63}\%$ & 52.70$\%$ & 52.45$\%$ & 50.29$\%$& 49.66$\%$ & 2.02s \\
                 &ten-step ATLAS & 82.94$\%$& 81.06$\%$ & $\boldsymbol{53.45}\%$ & $\boldsymbol{53.10}\%$ & $\boldsymbol{51.51}\%$ & $\boldsymbol{50.16}\%$  &1.88s\\
				\hline \hline  \TBstrut \TBstrut
				
				\parbox[t]{2mm}{\multirow{8}{*}{\rotatebox[origin=c]{90}{CIFAR-100}}}
				& zero-step JAC & 47.96$\%$& 44.67$\%$ & 15.39$\%$ & 15.18$\%$ & 14.35$\%$ & -  & 0.47s \\
				\cline{2-10}
				& one-step ADV & 48.88$\%$& 45.59$\%$  & 19.89$\%$ & 19.72$\%$ & 19.04$\%$ & - & 0.25s \\
                & one-step TRADES & $\boldsymbol{62.58}\%$ & $\boldsymbol{57.64}\%$ & 16.93$\%$ & 16.21$\%$ & 14.83$\%$  & -   & 0.53s \\
                & one-step ATLAS   & 60.98$\%$ & 57.04$\%$ & $\boldsymbol{26.30}\%$& $\boldsymbol{25.94}\%$ & $\boldsymbol{26.28}\%$ & -  & 0.73s  \\
				\cline{2-10}
				 & ten-step ADV & $\boldsymbol{60.72}\%$ & $\boldsymbol{58.15}\%$& 27.71$\%$ & 27.43$\%$ & 26.83$\%$& -  &1.37s\\
                 & ten-step TRADES & 59.76$\%$& 57.78$\%$  & 26.27$\%$ & 26.10$\%$ & 23.86$\%$ & - & 2.00s \\
                 & ten-step ATLAS & 59.62$\%$&57.83$\%$ & $\boldsymbol{29.12}\%$ & $\boldsymbol{28.89}\%$ & $\boldsymbol{27.77}\%$ &  -  & 1.87s\\
				\bottomrule
			\end{tabular}}
	\end{table*}
\textbf{Results} We first consider the challenging CIFAR-10 dataset. One-step ATLAS outperforms other one-step methods in all strong white box attacks. Even when compared with methods trained with 10 steps PGD, the performance gap between one-step ATLAS and multi-step ADV under the strongest Multi-T attack is below $1.5\%$. Although adding local and global penalties result in extra computational time for one-step ATLAS (0.73s) than one-step ADV (0.25s), the fact that ATLAS achieves comparable high robust accuracy to that of multi-step ADV (1.38s) with weak adversaries still allows a roughly 50$\%$ speed-up in training. For a fair comparison, we also evaluate an ADV model with 5 PGD steps. The 5-step ADV models requires slightly more time (0.75s per batch) but is less robust than one-step ATLAS by reaching $44.93\%$ accuracy under the Multi-T attack. When trained on strong adversaries, multi-step ATLAS wins over the other two multi-step methods in all strong white box attacks. Similar performances are observed on MNIST and CIFAR-100. We provided ablation and other comparison studies in the appendix.
\vspace{-10pt}
\section{Discussion}
\vspace{-10pt}
We have adopted a different perspective and proposed ATLAS for constructing losses that allow more effective use of adversaries. ATLAS can be used as an initialisation technique for other complicated tasks. For instance, model trained with ATLAS can be employed as the starting point for layer-wise adversarial training in improving certified robustness.

\bibliography{iclr2021_conference}
\bibliographystyle{iclr2021_conference}
\newpage
\appendix
\section*{A. Detailed analyses of ATLAS}
We give more details and thorough analytical reasonings for each component of ATLAS.
\subsection*{A.1. ATLAS-local}
Many existing approaches for increasing robust ball radius are computationally expensive. In this work, we adopt the standard Jacobian approach to meet our local component requirements based on the fact that an efficient estimation algorithm for Jacobian has been developed in \citet{Jac}. We briefly introduce how Jacobian can be used to increase local robustness.

Given a local point $\boldsymbol{x}'$, let the correct class label be $c = y_{\boldsymbol{x}}$. Then for any $c'\neq c$, the boundary hyper-surface separating classes $c$ and $c'$ consists of points $\boldsymbol{x}^b$ satisfying   
\begin{equation}\label{boundary}
    f_{c}(\boldsymbol{x}^b; \boldsymbol{\theta}) - f_{c'}(\boldsymbol{x}^b;\boldsymbol{\theta}) = 0.
\end{equation}

Applying the standard formula for computing the distance between a point and a hyper-plane \footnote{in this case, the point should be $\boldsymbol{x}'$ and the hyper-plane is tangent to the boundary hyper-surface.}, we get the first order approximation distance $d_{c'}$ of $\boldsymbol{x}'$ to the boundary hyper-surface in equation~(\ref{boundary}) under the $l_2$ norm as 
\begin{equation}
    d_{c'} = \frac{|f_{c}(\boldsymbol{x}'; \boldsymbol{\theta}) - f_{c'}(\boldsymbol{x}';\boldsymbol{\theta})|}{\Vert \nabla_{\boldsymbol{x}'}f_{c}(\boldsymbol{x}'; \boldsymbol{\theta}) - \nabla_{\boldsymbol{x}'}f_{c'}(\boldsymbol{x}'; \boldsymbol{\theta})\Vert_2}.
\end{equation}
Since the above equation holds true for any $c'$, we conclude the model is robust on an $l_2$ norm ball centered at $\boldsymbol{x}'$ with radius $d \coloneqq \min_{c'\neq c}d_{c'}$. To maximize the radius $d$, we borrow the following proposition from \citet{jac2018jakubovitz}, which introduced a Jacobian regularizer to the natural loss, to provide a lower bound for $d$. 
\begin{restatable}{prop}{jacprop}
Assume the model is making the correct prediction $c = y_{\boldsymbol{x}}$ for $\boldsymbol{x}'$ and the distance metric is measured via $l_2$ norm. The first order approximation of the minimum perturbation $d$ that is required to find an adversary example is lower bounded by
\begin{equation}
    d \geq \frac{1}{\sqrt{2}\Vert J(\boldsymbol{x}')\Vert_{F}}\min_{c'\neq c}|f_{c}(\boldsymbol{x}'; \boldsymbol{\theta}) - f_{c'}(\boldsymbol{x}';\boldsymbol{\theta})|,
\end{equation}
where $J(\boldsymbol{x}') = \nabla_{\boldsymbol{x}'}f(\boldsymbol{x}';\boldsymbol{\theta})$ is the Jacobian matrix computed at $\boldsymbol{x}'$ and $\Vert\cdot\Vert_{F}$ is the Frobenius norm.
\end{restatable}

For a larger distance $d$, we need to both increase the value of $|f_{c}(\boldsymbol{x}'; \boldsymbol{\theta}) - f_{c'}(\boldsymbol{x}';\boldsymbol{\theta})|$ and decrease the Frobenius norm of $J(\boldsymbol{x}')$. The first term can be easily taken care of by using a cross-entropy loss on $\boldsymbol{x}'$ while for the second term, we can include a cheap approximation of $\Vert J(\boldsymbol{x}')\Vert^2_{F}$, denoted by $\Vert J(\boldsymbol{x}')\Vert_{F}^{approx}$ as a penalty term in our loss. Using the idea of random projection, \cite{Jac} shows theoretically and empirically that $\Vert J(\boldsymbol{x}')\Vert_{F}^{approx}$ can be estimated with high quality by one backward pass regardless the total number of classes $C$. 

To combine the above analyses, for the local robustness component, we introduce the following loss,
\begin{equation}\label{loca-loss}
    \ell_{\text{local}}(\boldsymbol{x}') = \ell^{CE}(\boldsymbol{x}') + \alpha \cdot \Vert J(\boldsymbol{x}')\Vert_{F}^{approx},
\end{equation}
where $\alpha$ is a positive scalar. It is worth noting that our local loss shown in equation~(\ref{loca-loss}) is different from that of \citep{jac2018jakubovitz,Jac,ross2018improving} as the loss is evaluated at an adversary $\boldsymbol{x}'$ instead of the image $\boldsymbol{x}$. In the fast robust training setting, $\boldsymbol{x}'$ is a weak adversary obtained through one-step FGSM.

\subsection*{A.2. ATLAS-global}
In terms of the global component, we combine local balls together in a controlled way to further improve robustness. We first make the following observation.
\begin{restatable}{prop}{proppc}
\label{prop-pc}
Consider a ball $\mathcal{B}_{\epsilon}(\boldsymbol{x})$ at an arbitrary point $\boldsymbol{x}\in \mathcal{X}$. For any $\boldsymbol{x}^{\ast}\in\mathcal{B}_{\epsilon}(\boldsymbol{x})$, we define $\breve{W}^{\boldsymbol{x}^{\ast}}$ and $\breve{\boldsymbol{x}}^{\ast}$ accordingly, as in equation~(\ref{w-def}). Let $u$ be a constant such that, for all possible $\breve{W}^{\boldsymbol{x}^{\ast}}$, the following is satisfied $\Vert \breve{W}^{\boldsymbol{x}^{\ast}}\Vert_{F} \leq u$.
Assume $\boldsymbol{x}^{1}$ and $\boldsymbol{x}^{2}$ are arbitrary chosen points in $\mathcal{B}_{\epsilon}(\boldsymbol{x})$. Then for any
\begin{equation}
    \boldsymbol{x}_{\eta}  = \eta \cdot  \boldsymbol{x}^{1} + (1-\eta)\cdot \boldsymbol{x}^{2},
\end{equation}
we have 
\begin{equation}
 \boldsymbol{d}\breve{W}^{\boldsymbol{x}_{\eta}}\breve{\boldsymbol{x}}_{\eta}\geq\frac{1}{2}(\boldsymbol{d}\breve{W}^{\boldsymbol{x}^1}\breve{\boldsymbol{x}}^{1} +\boldsymbol{d}\breve{W}^{\boldsymbol{x}^2}\breve{\boldsymbol{x}}^{2}) - u\cdot L  ,
\end{equation}
where $L = \sqrt{2}(2\Vert \breve{\boldsymbol{x}}_{\eta}\Vert_2 +\frac{1}{2}\Vert \breve{\boldsymbol{x}}^{1} - \breve{\boldsymbol{x}}^{2}\Vert_2)$ is a constant. 
\end{restatable}

We now consider the combination of local robust balls. At each loss computation, we have two points: the natural image $\boldsymbol{x}$ and the adversary $\boldsymbol{x}'$. Assume the model makes the correct prediction at the natural image and, after the local component loss, at $\boldsymbol{x}'$, so we have two non-empty local robust balls $\mathcal{B}_{\gamma}(\boldsymbol{x})$ and $\mathcal{B}_{\gamma'}(\boldsymbol{x}')$. We further assume these two balls are disjoint for the sake of simplicity. The same underlying idea should work for more general cases. The goal is that, with the global regularization step, points that are misclassified even after the local step can be correctly predicted. 

To do so, given that local robust balls $\mathcal{B}_{\gamma}(\boldsymbol{x})$ and $\mathcal{B}_{\gamma'}(\boldsymbol{x}')$ are disjoint, we assume there exists an adversary point $\boldsymbol{x}^{\ast} \in \mathcal{B}_{\epsilon}(\boldsymbol{x})$ satisfying
\begin{equation}
    \boldsymbol{x}^{\ast} = \eta^{\ast} \cdot  \boldsymbol{x} + (1-\eta^{\ast})\cdot \boldsymbol{x}',
\end{equation}
where $\eta^{\ast}\in (0,1)$ is a constant depending on $\mathcal{B}_{\gamma}(\boldsymbol{x})$ and $\mathcal{B}_{\gamma'}(\boldsymbol{x}')$. Then a direct application of Proposition~\ref{prop-pc} gives
\begin{equation}
\begin{split}
     \boldsymbol{d}\breve{W}^{\boldsymbol{x}^{\ast}}\breve{\boldsymbol{x}}^{\ast}&\geq\frac{1}{2}(\boldsymbol{d}\breve{W}^{\boldsymbol{x}}\breve{\boldsymbol{x}} +\boldsymbol{d}\breve{W}^{\boldsymbol{x}'}\breve{\boldsymbol{x}}') - u\cdot L\\
      &= \frac{1}{2}\boldsymbol{d}(\breve{W}^{\boldsymbol{x}}\breve{\boldsymbol{x}}-\breve{W}^{\boldsymbol{x}'}\breve{\boldsymbol{x}}') + \boldsymbol{d}\breve{W}^{\boldsymbol{x}'}\breve{\boldsymbol{x}}' - u\cdot L
\end{split}
\end{equation}
where $u$ and $L$ are defined accordingly. 

To make $\boldsymbol{x}^{\ast}$ no longer an adversary (that is, to encourage $ \boldsymbol{d}\breve{W}^{\boldsymbol{x}^{\ast}}\breve{\boldsymbol{x}}^{\ast}>0$) after global combination, we should make the lower bound on the right as high as possible. 
Since both $\boldsymbol{d}\breve{W}^{\boldsymbol{x}}\breve{\boldsymbol{x}}$ and $\boldsymbol{d}\breve{W}^{\boldsymbol{x}'}\breve{\boldsymbol{x}}'$ have the same max value theoretically, we can increase their sum by first increasing the value of $\boldsymbol{d}\breve{W}^{\boldsymbol{x}'}\breve{\boldsymbol{x}}'$ and then decrease the distance between $\boldsymbol{d}\breve{W}^{\boldsymbol{x}}\breve{\boldsymbol{x}}$ and $\boldsymbol{d}\breve{W}^{\boldsymbol{x}'}\breve{\boldsymbol{x}}'$. We recall that during the local robustness step, we have introduced cross-entropy loss at $\boldsymbol{x}'$, which directly maximizes the value of $\boldsymbol{d}\breve{W}^{\boldsymbol{x}'}\breve{\boldsymbol{x}}'$. In addition, enlarging local robust balls often leads to a decreased value of $u$. Given that $\boldsymbol{d}\breve{W}^{\boldsymbol{x}'}\breve{\boldsymbol{x}}'$ and $u$ are already optimized, a sound global step is to minimise the distance between $\boldsymbol{d}\breve{W}^{\boldsymbol{x}}\breve{\boldsymbol{x}}$ and $\boldsymbol{d}\breve{W}^{\boldsymbol{x}'}\breve{\boldsymbol{x}}'$. We note that, for the distance to be zero, it is sufficient to have the relative differences $f_2(\boldsymbol{x}';\boldsymbol{\theta})-f_1(\boldsymbol{x}';\boldsymbol{\theta})$ and $f_2(\boldsymbol{x};\boldsymbol{\theta})-f_1(\boldsymbol{x};\boldsymbol{\theta})$ to be equal rather than requiring an equality between logits $f(\boldsymbol{x};\boldsymbol{\theta})$ and $f(\boldsymbol{x}';\boldsymbol{\theta})$. To account for this fact, we use prediction probabilities instead. The global regularization step requires minimizing the prediction probability difference $p(\boldsymbol{x};\boldsymbol{\theta})-p(\boldsymbol{x}';\boldsymbol{\theta})$. Depending on the problem at hand, one can choose an appropriate regularizer for increasing local robust balls while selecting a suitable metric for penalizing the prediction probability difference.
To penalize the prediction probability difference, a natural and frequently used choice is Kullback–Leibler (KL) distance. We thus formulate the global loss as
\begin{equation}
    \ell_{\text{global}}(\boldsymbol{x}') = \beta \cdot \text{KL}(p(\boldsymbol{x}';\boldsymbol{\theta})\|p(\boldsymbol{x};\boldsymbol{\theta})),
\end{equation}
where $\beta$ is a positive constant. 

\subsection*{A.3. Final Loss}
Integrating the above analyses, we propose the following as the final loss for ATLAS. Given $\alpha>0$ and $\beta >0$, the loss is formulated as
\begin{equation}
    \ell_{\text{ATLAS}} = \frac{1}{|\mathcal{X}|}\sum_{\boldsymbol{x}\in\mathcal{X}} \underbrace{\ell^{\text{CE}}(\boldsymbol{x}') + \alpha \cdot \Vert J(\boldsymbol{x}')\Vert_{F}^{approx}}_{\text{local}} + \underbrace{\beta \cdot \text{KL}(p(\boldsymbol{x}';\boldsymbol{\theta})\|p(\boldsymbol{x};\boldsymbol{\theta}))}_{\text{global}}, \quad \boldsymbol{x}'\in \mathcal{B}_{\epsilon}(\boldsymbol{x}).
\end{equation}

We mention that although our analyses are carried out in $l_2$ norm, our results are easily generalizable to other norms. To ensure the model makes correct predictions at natural images $\boldsymbol{x}\in\mathcal{X}$ and for effective patch combination, we adopt an adaptive value scheme for the pre-determined perturbation epsilon. Specifically, we start by setting $\epsilon=0$ and gradually increases its value to the required number over epochs during the initial stage of training. Assume the final value of $\epsilon$ is $v$, to be consistent with the magnitude of $\epsilon$, we replace $\alpha$ and $\beta$ as $\frac{\epsilon}{v}\cdot \alpha$ and $\frac{\epsilon}{v}\cdot \beta$ accordingly. 

\subsection*{A.4. Proof for Proposition 2}
\proppc*
\begin{proof}
The proof is straightforward. To simplify the notation, we define $s^1 \coloneqq \boldsymbol{d}\breve{W}^{\boldsymbol{x}^1}\breve{\boldsymbol{x}}^1 $, $s^2 \coloneqq \boldsymbol{d}\breve{W}^{\boldsymbol{x}^2}\breve{\boldsymbol{x}}^2$ and $s_{\eta} \coloneqq \boldsymbol{d}\breve{W}^{\boldsymbol{x}_{\eta}}\breve{\boldsymbol{x}}_{\eta}$. Also, let $l$ be the lower bound $\boldsymbol{d}\breve{W}^{\boldsymbol{x}^{\ast}}\breve{\boldsymbol{x}}^{\ast} \geq l$ for any $\boldsymbol{x}^{\ast}\in\mathcal{B}_{\epsilon}(\boldsymbol{x})$. Such $l$ exits because $\mathcal{B}_{\epsilon}(\boldsymbol{x})$ is a closed and bounded ball and $f$ is continuous. We first note that 
\begin{equation*}
     \vert s^1 - l \vert - \vert s^1 - s_{\eta}\vert \leq \vert s_{\eta} - l\vert .
\end{equation*}
Since $ s^1 - l >0$ and $ s_{\eta} - l\geq 0$, we can remove the absolute sign and have 
\begin{equation}\label{bi-ineq}
     s^1  - \vert s^1 - s_{\eta}\vert \leq s_{\eta}.
\end{equation}
The term $\vert s^1-s_{\eta}\vert$ can be upper bounded as 
\begin{equation}\label{bounds}
   \begin{split}
    \vert s^1-s_{\eta}\vert &= \vert\boldsymbol{d}\breve{W}^{\boldsymbol{x}^1}\breve{\boldsymbol{x}}^1 -\boldsymbol{d}\breve{W}^{\boldsymbol{x}_{\eta}}\breve{\boldsymbol{x}}_{\eta} \vert \\
    & \leq \vert (\boldsymbol{d}\breve{W}^{\boldsymbol{x}^1}- \boldsymbol{d}\breve{W}^{\boldsymbol{x}_{\eta}})\breve{\boldsymbol{x}}_{\eta}\vert + \vert\boldsymbol{d}\breve{W}^{\boldsymbol{x}^1}(\breve{\boldsymbol{x}}^1-\breve{\boldsymbol{x}}_{\eta}) \vert \\
    & \leq 2u \Vert \boldsymbol{d}\Vert_2\Vert \breve{\boldsymbol{x}}_{\eta}\Vert_2 +u (1-\eta) \Vert \boldsymbol{d}\Vert_2\Vert \breve{\boldsymbol{x}}^1 - \breve{\boldsymbol{x}}^{2}\Vert_2.
\end{split} 
\end{equation}

We further denote $m =2u \Vert \boldsymbol{d}\Vert_2\Vert \breve{\boldsymbol{x}}_{\eta}\Vert_2 $ and $v =u\Vert\boldsymbol{d}\Vert_2\Vert \breve{\boldsymbol{x}}^1 - \breve{\boldsymbol{x}}^{2}\Vert_2 $. By replace $ \vert s^1-s_{\eta}\vert$ with its upper bounds equation~(\ref{bounds}), it follows that  
\begin{equation}\label{bi-f1}
     s^1  - m - (1-\eta)v \leq s_{\eta} .
\end{equation}
Similarly for $s^2$, we have 
\begin{equation}\label{bi-f2}
     s^2   - m - \eta v \leq s_{\eta}.
\end{equation}
 Finally, combining equation~(\ref{bi-f1}) and equation~(\ref{bi-f2}), we get 
 \begin{equation}
     \frac{1}{2} (s^1+s^2) -m - \frac{1}{2}v \leq s_{\eta},
 \end{equation}
the desired inequality.
\end{proof}

\section*{B. More Experimental Details and Analyses}
We give more experimental details and analyses in the following.

\textbf{Training } To encourage efficient robust training, instead of a fixed learning rate scheduler that decreases the learning rate at pre-specified epoch numbers, we randomly choose a subset of the data to be the validation set and then use its robust accuracy to guide the learning rate adjustment and to terminate the training process. For MNIST, 20-step PGD is applied for computing robust accuracy on the validation set while for CIFAR-10 and CIFAR-100, 10-step PGD is used. In addition, we gradually increase epsilon value from 0 to $0.3$ or $8/255$ for the first 15 epochs. We use the SGD optmizer for the training. 

We rely on the robust accuracy on a validation dataset to guide the training process. To be more specific, we first randomly select 5000 images from the training dataset to form a validation set before training starts. During training, after each epoch, we run a multi-step PGD attack to evaluate the model's robust accuracy on the validation set. If the validation robust accuracy does not improve for a fixed number of consecutive epochs (we refer to the number as plateau epoch number), we decrease the learning rate by five. If the robust accuracy does not improve for ten consecutive epochs, we terminate the training process.

MNIST is trained on a 4-layer CNN, which consists of 2 convolutional layers followed by 2 fully connected layers. We use SGD optimizer with a starting learning rate of 0.01 on MNIST. Batch size is set to be 128. After each epoch, we apply a 20-step PGD attack with step size 0.01 to determine validation robust accuracy. Similarly, when generating strong adversaries for multi-step methods, 20-step PGD attack with step size 0.01 is used.    

CIFAR-10 and CIFAR-100 are trained on Wide-ResNet-28-8 \citep{wideres}. On both CIFAR-10 and CIFAR-100, we employ the SGD optimizer with a starting learning rate of 0.1, a momentum of 0.9 and a weight decay of $2e-4$. A batch size of 64 is used. In terms of computing validation robust accuracy and generating strong adversaries in multi-step cases, we apply 10-step PGD attack with step size 0.007. 

\textbf{Methods } For one-step methods, we follow the advice from \citet{wong2020fast}. We apply FGSM, combined with random initialization and a step size of 1.25$\epsilon$ to generate adversarial examples. To be consistent with their multi-step variants, we use cross-entropy loss for computing the gradient in FGSM for ADV and ATLAS while employ KL distance for TRADES. 

For each method, a range of parameters are tested and those that give both high clean accuracy and high validation robust accuracy are chosen. Results for all tested parameter choices are provided in Appendix F. In terms of numbers reported in Table \ref{table:mainTable}, for MNIST, zero-step JAC is trained with $\alpha_{\text{JAC}}=0.5$; one-step and multi-step TRADES are trained with $\beta_{\text{TRADES}}=1$; one-step ATLAS is trained with $\alpha=1e-05$ and $\beta=0.3$ while multi-step ATLAS is trained with $\alpha=5e-06$ and $\beta=0.2$. On the CIFAR-10 dataset, the following parameters are used for the numbers reported in Table \ref{table:mainTable}: zero-step JAC is trained with $\alpha_{\text{JAC}}=0.5$; one-step TRADES is trained with $\beta_{\text{TRADES}}=2$; multi-step TRADES is trained with $\beta_{\text{TRADES}}=4$; one-step ATLAS is trained with $\alpha=0.0002$ and $\beta=5$ and multi-step ATLAS is trained with $\alpha=0.0001$ and $\beta=5.0$. On the CIFAR-100 dataset, we used $\alpha_{\text{JAC}}=0.5$ for zero-step JAC; $\beta_{\text{TRADES}}=2$ for one-step TRADES; $\beta_{\text{TRADES}}=4$ for multi-step TRADES; $\alpha=0.0002$ and $\beta=5.0$ for one-step ATLAS and $\alpha=0.0001$ and $\beta=10.0$ for multi-step ATLAS. 

The ablation study on the effect of each component of ATLAS: the local part (ATLAS-l, by setting $\beta=0$) and the global part (ATLAS-g, by setting $\alpha=0$) is given in Appendix C and the same is for a detailed comparison between TRADES and ATLAS-g. In addition, we have include in Appendix D.1 an experiment to demonstrate ATLAS's effective use of weak adversaries. We compare ATLAS against a loss which always treats the natural image $\boldsymbol{x}$, instead of a weak adversary $\boldsymbol{x}'$, as the center point for the local component. In the words, the loss takes the form of combining TRADES and zero-step JAC. 

\textbf{Robustness Evaluation } Random initialization is applied in all attacks. Untargeted and Multi-targeted attacks are implemented by following the descriptions given in \citep{LLR}. For these two attacks, we consider an attack is successful if an adversary is found after a gradient update at any point during the optimization procedure. Untargeted attack uses the loss $f_u(\boldsymbol{x};\boldsymbol{\theta})-f_{c}(\boldsymbol{x};\boldsymbol{\theta})$, where $c$ is the correct class and $u = \argmax_{i\neq c}f_{i}(\boldsymbol{x})$. We allow $u$ to change during each gradient step. Regarding Multi-targeted attack, we perform attack with the loss $f_i(\boldsymbol{x};\boldsymbol{\theta})-f_{c}(\boldsymbol{x};\boldsymbol{\theta})$ for all $i\in \mathcal{C}$ such that $i\neq c$. On the big Wide-ResNet model, we run 10 restarts with 100 steps for Untargeted attack and 5 restarts with 20 steps for Multi-target attack. We reduce the number of restarts and steps for the Multi-target attack to account for the fact that each incorrect class needs to be tested. Due to the large number of classes, Multi-target is not performed on CIFAR-100. On the other hand, we perform 20 restarts with 50 steps for both attacks on the small CNN model. For each method, attack results are reported for a model at the epoch that gives the best validation robust accuracy during the training. On the challenging datasets CIFAR-10 and CIFAR-100, two more robust evaluations are carried out to obtain a more comprehensive understanding of models' robustness performance. Specifically, we apply an hard-label attack, RayS \citep{rays} and an ensemble of diverse parameter-free attacks, AutoAttack \citep{AA}. Models' robustness performance under these two evaluations are reported in Appendix D.2. 

\textbf{Results Analyses } We give more details on results. We mention that with a learning rate scheduler, which is guided by the robust accuracy on the validation set, all models achieved their best performance around 30 epochs, except 12 epochs for zero-step JAC on MNIST. In terms of robustness performance, slight performance improvements for all methods and narrower performance gaps will be observed if all models are trained for 70 epochs with a fixed learning rate scheduler on MNIST. On the other hand, on CIFAR-10 and CIFAR-100, we found all methods perform better with the guided learning rate scheduler. Since we are interested in efficient robust learning, we use the guided learning rate scheduler for all experiments. 

\textbf{More on CIFAR-10 } We provide more analyses for the challenging dataset CIFAR-10. For zero-step Jac, we observe that to achieve a roughly $30\%$ robust accuracy, nominal accuracy is dropped to below $65\%$. Directly including a Jacobian penalty at natural images could thus lead to over-regularization issues. This may also be the reason why the method is used as a post-processing technique in \citep{jac2018jakubovitz}. On the other hand, by using local adversaries instead of natural images, ATLAS-l manages to obtain robust accuracy without sacrificing the nominal accuracy too much, which is demonstrated in the Appendix C. When TRADES is considered, we find that its effectiveness relies on strong adversaries: robust accuracy increases sharply from one-step case to multi-step case. Since TRADES computes loss at natural images and use them as the base distribution in the KL penalty, weak adversaries are not effectively used. 

In addition, we constantly find for one-step TRADES that, after a certain number of epochs, the percentage of the model making incorrect decisions on weak adversaries falls sharply, leading to a sudden drop of more than $10\%$ on validation robust accuracy. This phenomena is referred to as catastrophic overfitting \citep{wong2020fast, understanding2020}. We observe that when this behaviour occurs, the approximated value $\Vert J(\boldsymbol{x}')\Vert_{F}^{approx}$ increases steeply, which is consistent with what has been observed in \citet{understanding2020}. When dealing with catastrophic overfitting is the main concern, \citet{understanding2020} argue the key is to increase local linearity and they introduce a regularizer to maximize gradient alignment for points within the perturbation ball. In our case, the issue could be similarly resolved by increasing the coefficient $\alpha$ to encourage local linearity. However, since the goal of this project is to achieve high robust accuracy fast, we do not restrict ourselves to models without catastrophic overfitting only, which are likely to compensate robust accuracy for stability. Instead, we use early termination and consider a wider range of models. We mention that the same phenomenon is observed on one-step ADV and on one-step ATLAS when $\alpha$ is small but less frequently. More discussions on catastrophic overfitting can be found in Appendix E. 

\subsection*{B.1. Plateau epoch number and FGSM step-size}
Due to the simplicity of the MNIST dataset and similar behaviour on both CIFAR-10 and CIFAR-100, we determine the plateau epoch number and the FGSM step-size by experimenting with one-step ADV on CIFAR-10.

\begin{figure}[h]
 \begin{center}
    \includegraphics[width=0.45\textwidth]{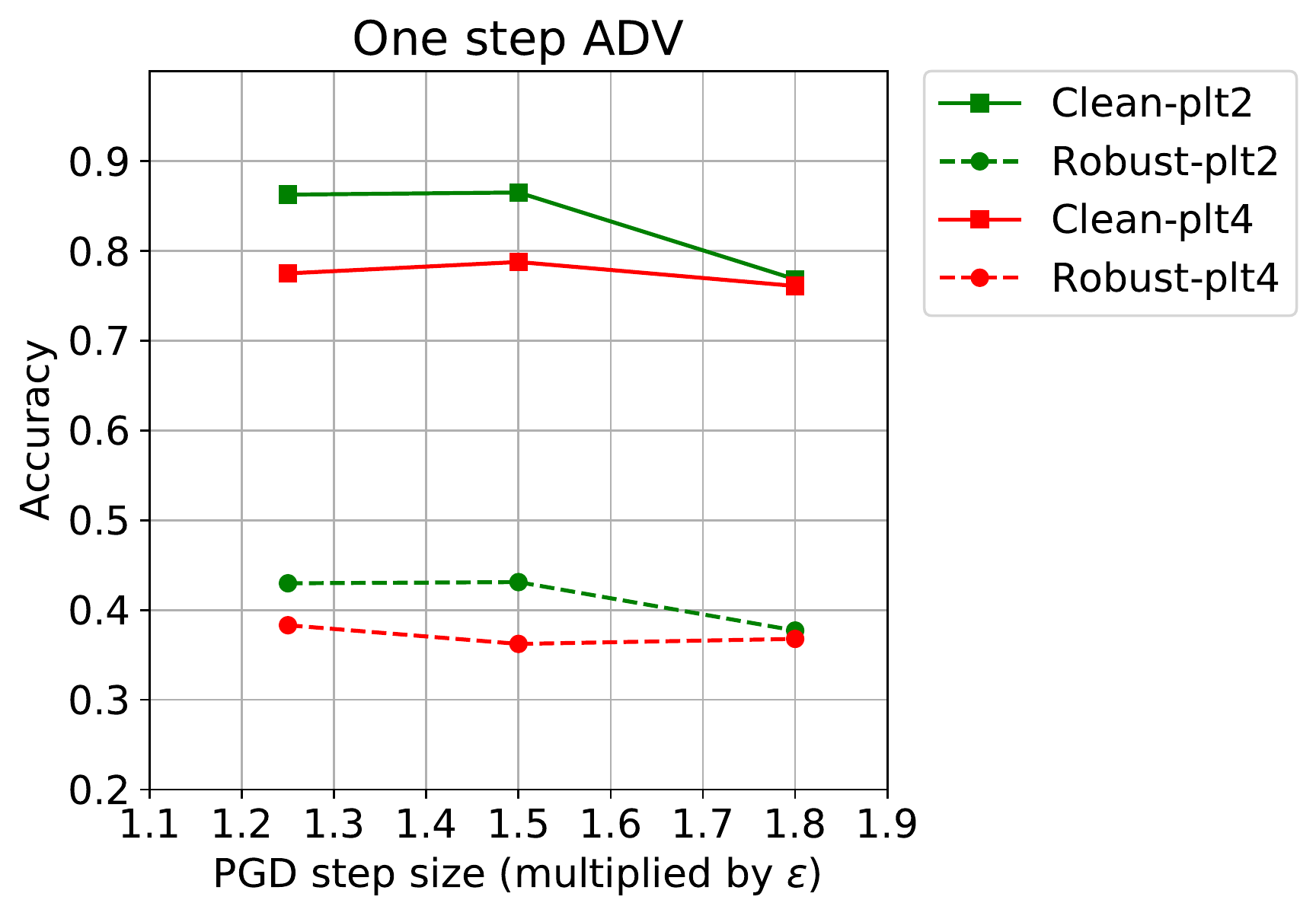}
  \end{center}
\caption{Clean and robust accuracy for one-step ADV at different step sizes}
\label{fig:onestepadv}
\end{figure}
We use strongest Multi-T attack for evaluating model's robust accuracy. In Figure \ref{fig:onestepadv}, the green lines show the results for plateau epochs to be 2 (plt2) while red lines are for plateau epochs to be 4 (plt4). In terms of step sizes, we tested 3 possible values: $1.25\epsilon$ (the suggested value by \citet{wong2020fast}), $1.5\epsilon$ and $1.8\epsilon$. It is clear to see that with plt2, models perform better in both nominal accuracy and robust accuracy at all three step sizes. We thus use plateau epoch number 2 to adjust the learning rate in all our experiments for comparable results. When step size is considered, both $1.25\epsilon$ and $1.5\epsilon$ give satisfactory results. Since other methods show more stable performances with the step size $1.25\epsilon$, we use it for generating weak adversaries in our experiments. 

\section*{C ATLAS: ablation studies}
We perform ablation studies for ATLAS. We consider the one-step setting. Recall that ATLAS consists of two components: a local component and a global component. We test the effect of the local component by setting $\beta=0$ to get \textbf{ATLAS-l}:
\begin{equation}
    \ell_{\text{ATLAS-l}} = \frac{1}{|\mathcal{X}|}\sum_{\boldsymbol{x}\in\mathcal{X}} \ell^{\text{CE}}(\boldsymbol{x}') + \alpha \cdot \mathcal{A} (J(\boldsymbol{x}')),\quad \boldsymbol{x}'\in \mathcal{B}_{\epsilon}(\boldsymbol{x}).
\end{equation}
and the global component by setting $\alpha=0$ to get \textbf{ATLAS-g}:
\begin{equation}
    \ell_{\text{ATLAS-g}} = \frac{1}{|\mathcal{X}|}\sum_{\boldsymbol{x}\in\mathcal{X}}\ell^{\text{CE}}(\boldsymbol{x}') + \beta \cdot \text{KL}(p(\boldsymbol{x}';\boldsymbol{\theta})\|p(\boldsymbol{x};\boldsymbol{\theta})), \quad \boldsymbol{x}'\in \mathcal{B}_{\epsilon}(\boldsymbol{x}).
\end{equation}
We apply the strongest Multi-T attack on CIFAR-10 and Un-T attack on CIFAR-100. Results are summarised in Table \ref{table:ablation}. When CIFAR-10 is considered, both ATLAS-l and ATLAS-g alone are effective in improving the model's robust accuracy. Combining the local and global components (one-step ATLAS) lead to a further robustness improvement. In terms of time per batch, apart from the cross-entropy loss term at the adversary in both ATLAS-l and ATLAS-g, ATLAS-l requires computing the gradient of a Jacobian approximation term and is computationally more expensive than ATLAS-g, which calculates the gradient of a KL penalty term instead. Furthermore, the fact that ATLAS-g is computationally cheaper than TRADES is because it uses cross-entropy loss to compute gradients for FGSM while TRADES uses KL distance. Similar performance is observed on CIFAR-100. 

\begin{table*}[h]
\centering
	\rtable{1.2}
			\caption{Robustness performance for CIFAR-10 and CIFAR-100 on Wide-Resnet 28-8. The higher the better. \label{table:ablation}}
			\resizebox{0.7\linewidth}{!}{
			\begin{tabular}{clcccc}
				\toprule
				& Model & \vtop{\hbox{\strut Clean}\hbox{\strut accuracy }} &\vtop{\hbox{\strut Un-T }\hbox{\strut attack}} & \vtop{\hbox{\strut Multi-T}\hbox{\strut attack}}   & \vtop{\hbox{\strut Time}\hbox{\strut per batch}} \\
				\midrule
				\parbox[t]{2mm}{\multirow{5}{*}{\rotatebox[origin=c]{90}{CIFAR-10}}}
				& one-step ADV & 86.24$\%$ &- &42.97$\%$  & 0.25s \\
				& one-step TRADES & $\boldsymbol{86.84}\%$  & -  & 38.14$\%$& 0.55s \\
                & one-step ATLAS-l & 85.24$\%$ &- & 44.46$\%$ & 0.60s  \\
                & one-step ATLAS-g & 82.07$\%$ &- &44.43$\%$ &  0.38s \\
                & one-step ATLAS  & 84.52$\%$ &- & $\boldsymbol{46.55}\%$&  0.73s  \\
				\hline \hline  \TBstrut \TBstrut
				
				\parbox[t]{2mm}{\multirow{5}{*}{\rotatebox[origin=c]{90}{CIFAR-100}}}
				& one-step ADV & 48.88$\%$& 19.04$\%$ & - & 0.25s \\
				& one-step TRADES & $\boldsymbol{62.58}\%$ & 14.83$\%$  & -   & 0.53s \\
				& one-step ATLAS-l & 58.46$\%$ & 24.73$\%$ & - & 0.60s\\
				& one-step ATLAS-g & 51.88$\%$ & 19.71$\%$ & - & 0.37s\\
                & one-step ATLAS   & 60.98$\%$ &  $\boldsymbol{26.28}\%$ & -  & 0.73s\\
				\bottomrule
			\end{tabular}}
	\end{table*}
	
\begin{figure}[h]
  \centering
   \begin{minipage}[t]{0.33\textwidth}
    \includegraphics[width=0.8\textwidth]{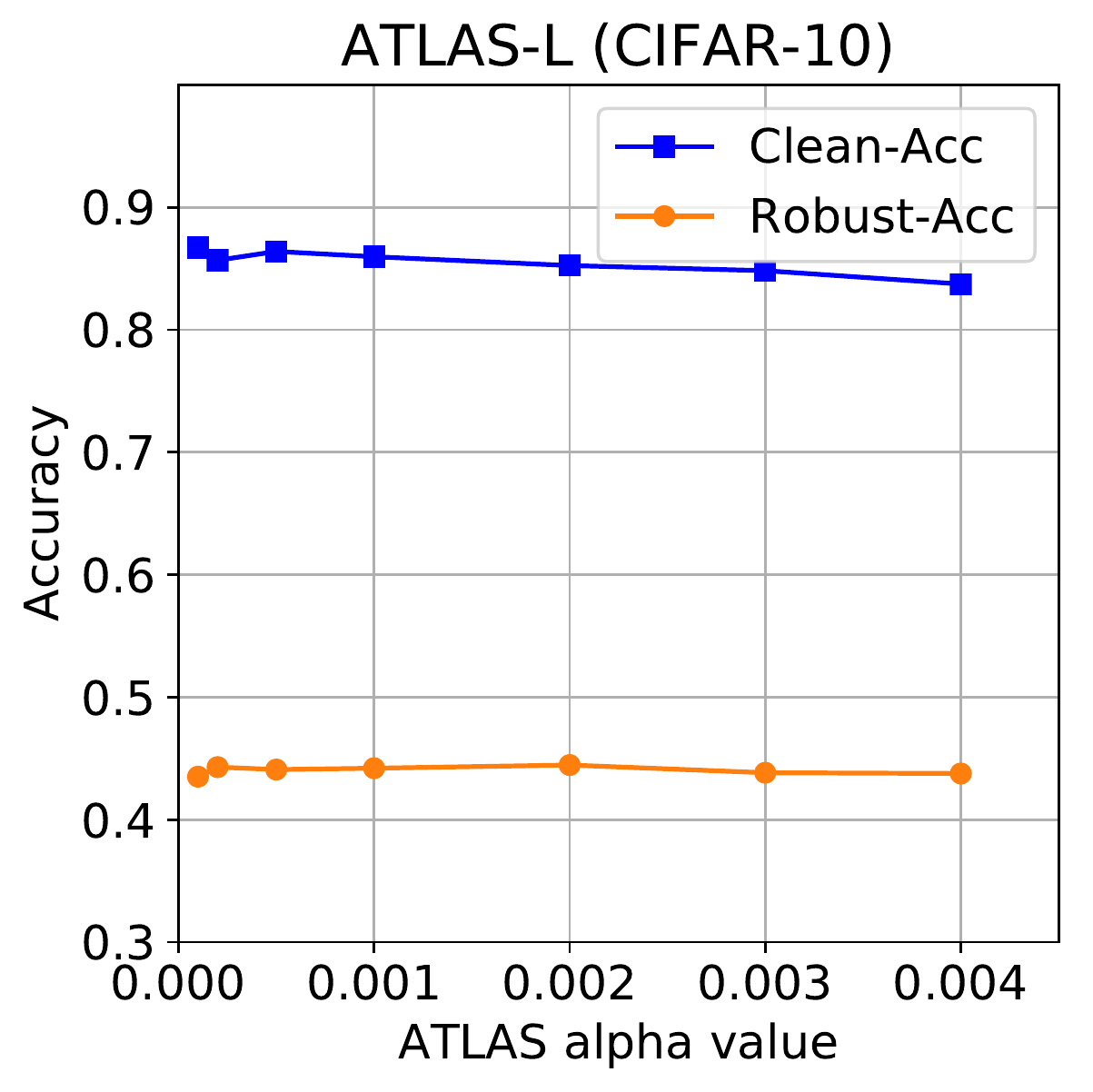}
    \caption{Clean and robust accuracy for ATLAS-l at different $\alpha$ on CIFAR-10}
    \label{fig:leap-l-c10}
  \end{minipage}
  \hfill
  \begin{minipage}[t]{0.6\textwidth}
    \includegraphics[width=0.8\textwidth]{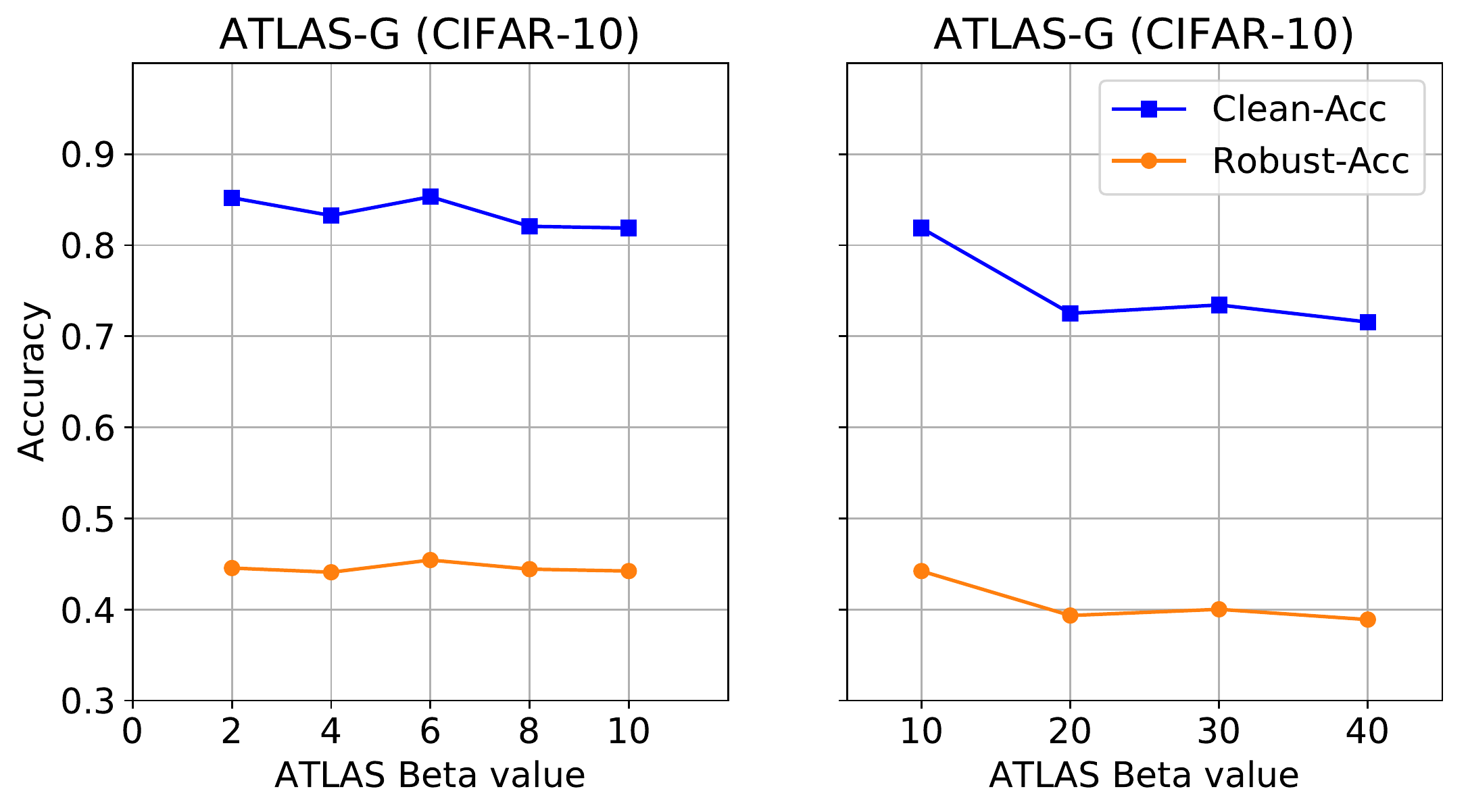}
    \caption{Clean and robust accuracy for ATLAS-l at different $\beta$ on CIFAR-10}
    \label{fig:leap-g-c10}
  \end{minipage}
\end{figure}

\begin{figure}[h]
  \centering
   \begin{minipage}[t]{0.33\textwidth}
    \includegraphics[width=0.8\textwidth]{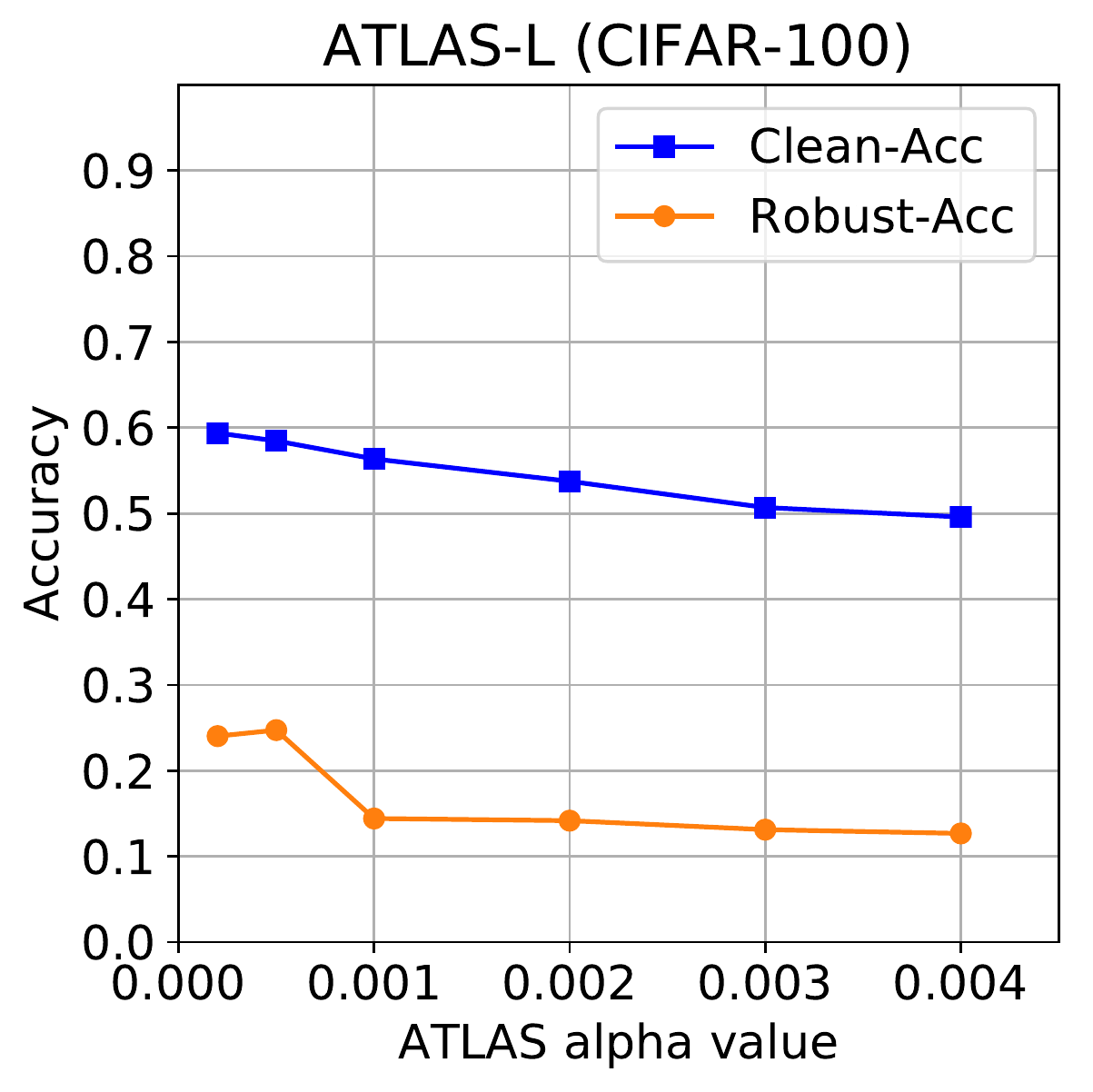}
    \caption{Performance of ATLAS-l at different $\alpha$ on CIFAR-100}
    \label{fig:leap-l-c100}
  \end{minipage}
  \hfill
  \begin{minipage}[t]{0.6\textwidth}
    \includegraphics[width=0.8\textwidth]{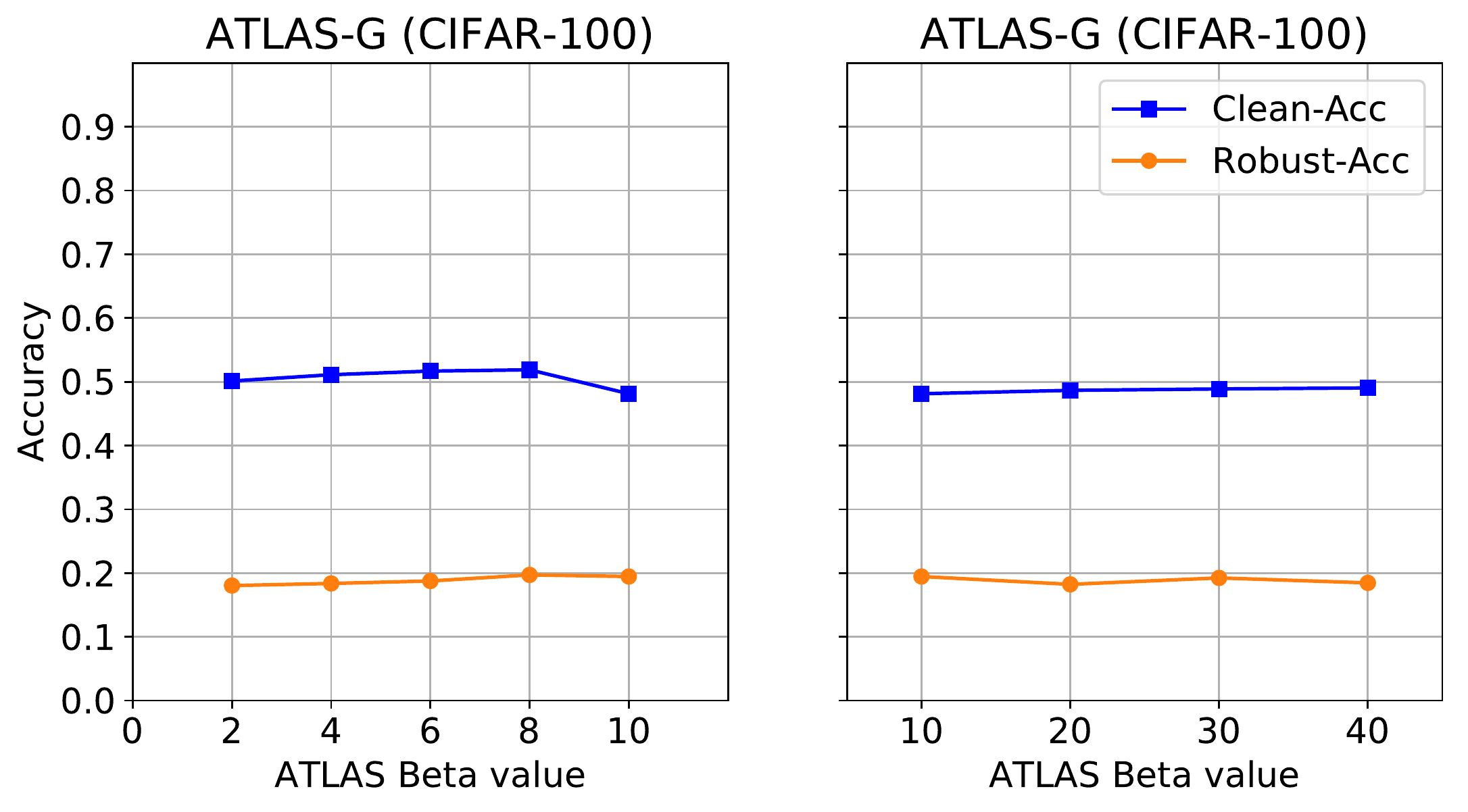}
    \caption{Clean and robust accuracy for ATLAS-l at different $\beta$ on CIFAR-100}
    \label{fig:leap-g-c100}
  \end{minipage}
\end{figure}

We show model's performance when trained with ATLAS-l and ATLAS-g at different parameter values. Results for CIFAR-10 are summarised in Figure \ref{fig:leap-l-c10} and \ref{fig:leap-g-c10} while results for CIFAR-100 are summarised in Figure \ref{fig:leap-l-c100} and \ref{fig:leap-g-c100}. For CIFAR-10, we see that the clean accuracy decreases with the increase of $\alpha$ value for ATLAS-l but robust accuracy retains at the similar level. In terms of ATLAS-g, the same trend is observed and when the value of $\beta$ is large, both clean and robust accuracy decline. On CIFAR-100, there is a big drop in robust accuracy for ATLAS-l when $\alpha$ rises while robust accuracy for ATLAS-g is relatively insensitive to the change of $\beta$ value.   

\subsection*{C.1 ATLAS-g vs. TRADES}
ATLAS-g and TRADES take similar forms. The only differences between these two methods are: firstly, ATLAS-g computes cross-entropy loss at the adversary while TRADES at the natural image; secondly, ATLAS-g uses cross-entropy loss to find a gradient in FGSM while TRADES employs KL distance. 

On CIFAR-10, it is clear that ATLAS-g outperforms TRADES on both clean and robust accuracy for all tested $\beta$ values. This observation supports the fact that ATLAS-g, by computing the loss at weak adversaries, allows a more effective use of them. Maximizing the use of weak adversaries is important in fast robust training setting. In terms of CIFAR-100, ATLAS-g outperforms TRADES on robust accuracy. However, TRADES achieves higher clean accuracy on small $\beta$ values. The regularization effect of ATLAS-g could be higher than TRADES. 
\begin{figure}[h]
  \centering
   \begin{minipage}[t]{0.35\textwidth}
    \includegraphics[width=0.8\textwidth]{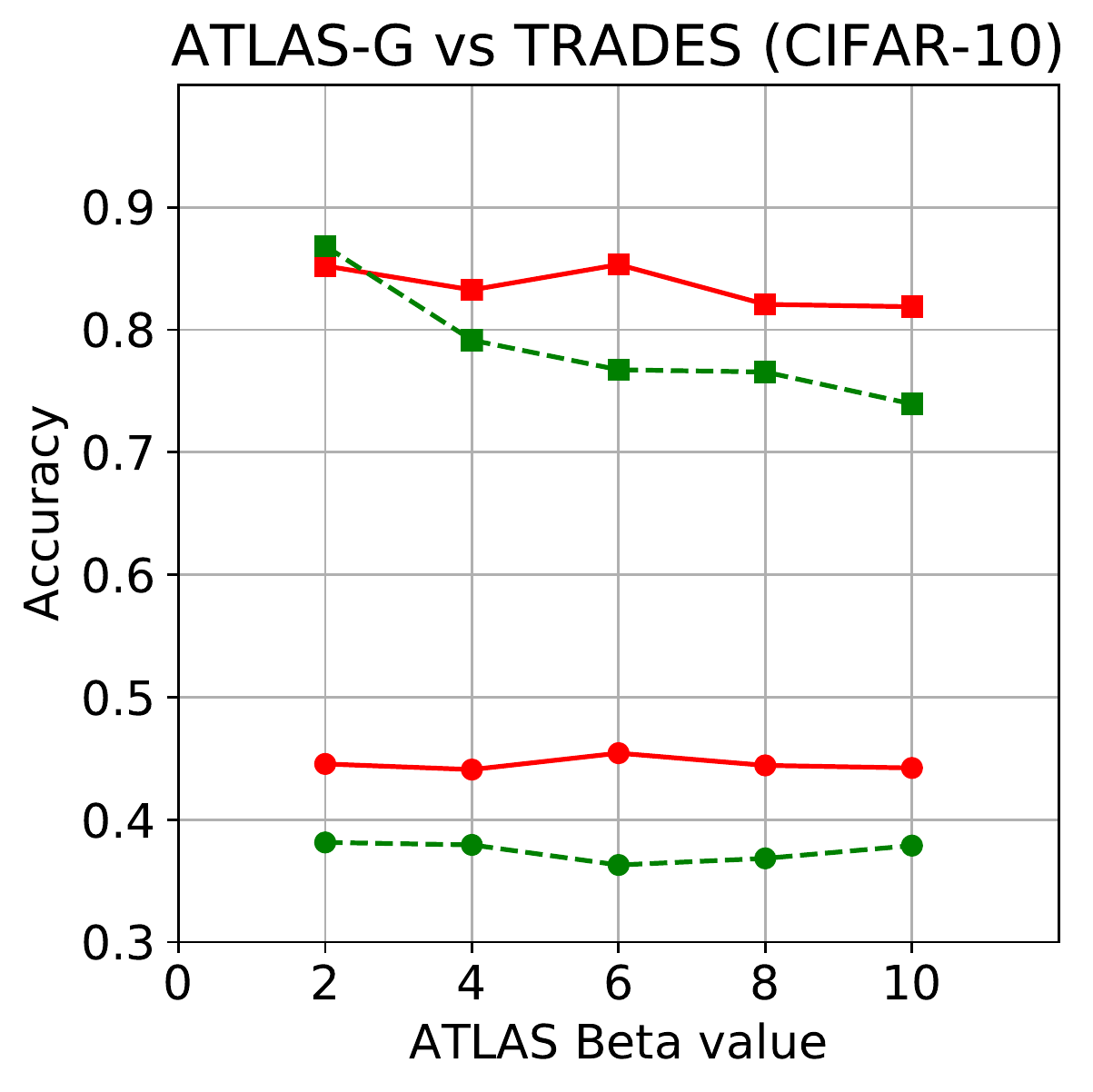}
    \caption{ATLAS-g vs. TRADES on CIFAR-10}
    \label{fig:leapgtrades-c10}
  \end{minipage}
  \hfill
  \begin{minipage}[t]{0.55\textwidth}
    \includegraphics[width=0.8\textwidth]{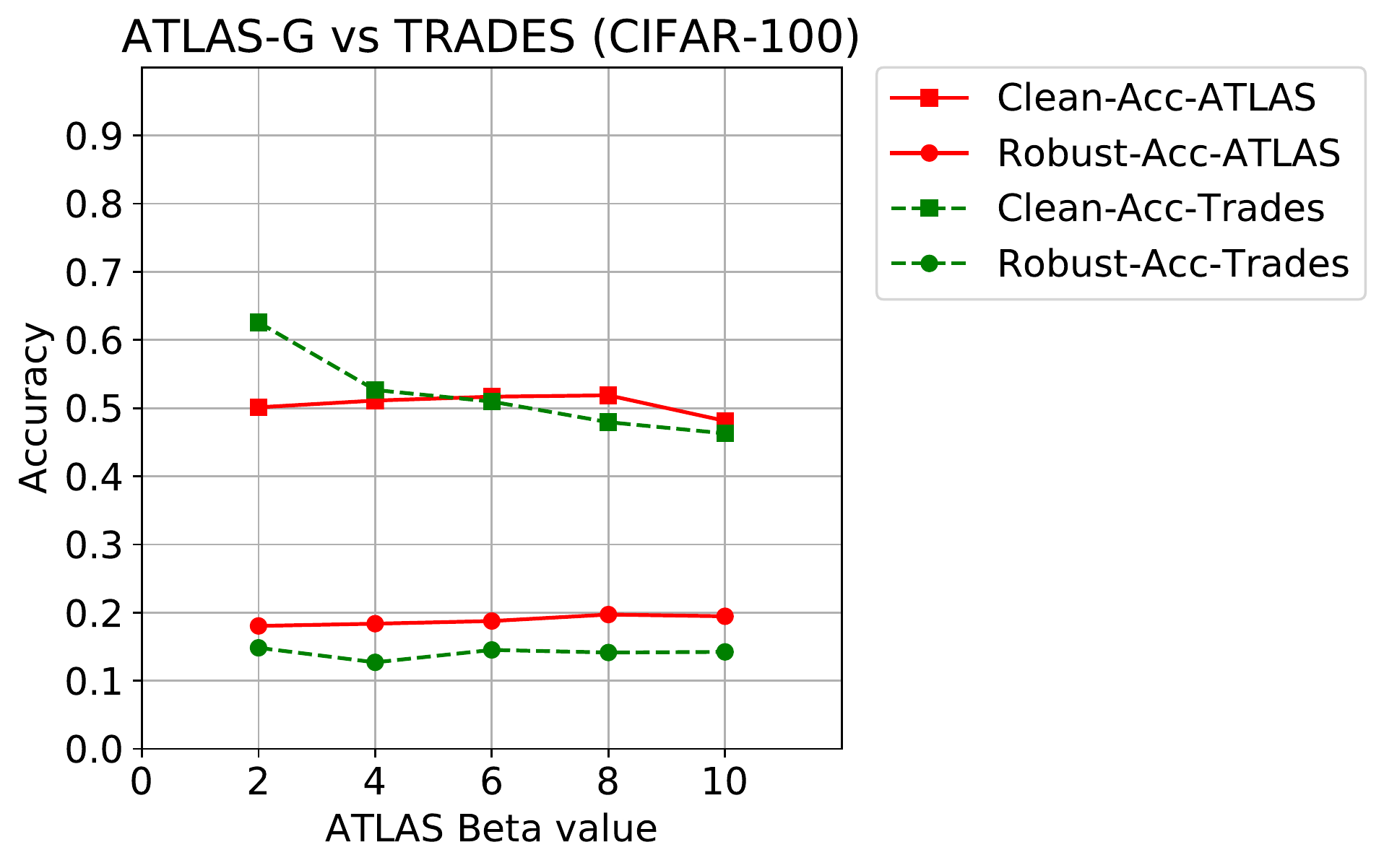}
    \caption{ATLAS-g vs. TRADES on CIFAR-100}
    \label{fig:leapgtrades-c100}
  \end{minipage}
\end{figure}

\section*{D Additional Experiments}
We compare ATLAS against another loss of similar form. Furthermore, we perform two extra attacks to obtain more comprehensive evaluations of all tested models. Due to the simplicity of MNIST, we focus on the challenging datasets CIFAR-10 and CIFAR-100 only. 

\section*{D.1 Additional Loss}
We study an additional loss to demonstrate ATLAS's effective use of weak adversaries. We recall that within our framework, the key to maximize the use of weak adversaries is to treat them as center points of local balls in the local component. We thus consider the following loss
\begin{equation*}
    \ell_{\text{TradesJac}} = \frac{1}{|\mathcal{X}|}\sum_{\boldsymbol{x}\in\mathcal{X}} \underbrace{\ell^{\text{CE}}(\boldsymbol{x}) + \alpha_{\text{tj}} \cdot \Vert J(\boldsymbol{x})\Vert_{F}^{approx}}_{\text{local}} + \underbrace{\beta_{\text{tj}} \cdot \text{KL}(p(\boldsymbol{x};\boldsymbol{\theta})\|p(\boldsymbol{x}';\boldsymbol{\theta}))}_{\text{global}}, \quad \boldsymbol{x}'\in \mathcal{B}_{\epsilon}(\boldsymbol{x}).
\end{equation*} 
We term the loss as TradesJac, as it can also been seen as a combination of the TRADES loss and the zero-step JAC loss. Adversaries are computed by using the gradient of the KL-distance. The main difference between TradesJac and ATLAS is in the local component, where TradesJac uses natural images as center points for local balls instead of weak adversaries. 

We train models using TradesJac with various sets of values for $\alpha_{\text{tj}}$ and $\beta_{\text{tj}}$ and reported results for the model with the best validation robust accuracy. Results can be found in Table \ref{table:add_loss}. On both CIFAR-10 and CIFAR-100, we have set $\alpha_{\text{tj}}=0.0002$ and $\beta_{\text{tj}}=0.5$. We have also include other one-step methods for easy comparison. It is clear to see that ATLAS outperforms TradesJac in all attacks, confirming that ATLAS is able to employ weak adversaries more effectively in achieving robust accuracy.  

\begin{table*}[h]
\centering
	\rtable{1.2}
			\caption{Robustness performance for one-step methods on CIFAR-10 and CIFAR-100. All models are trained with Wide-Resnet 28-8. The higher the better. \label{table:add_loss}}
			\resizebox{\linewidth}{!}{
			\begin{tabular}{clccccccccc}
				\toprule
				{} & {} & {}  & {}& \multicolumn{4}{c}{White box attack}  & \multicolumn{1}{c}{Time}\\
				\cmidrule(lr){5-8}\cmidrule(lr){9-9}
				& Model & \vtop{\hbox{\strut Clean}\hbox{\strut accuracy }} & \vtop{\hbox{\strut Black box}\hbox{\strut attack}}& \vtop{\hbox{\strut PGD-20}\hbox{\strut attack}} & \vtop{\hbox{\strut PGD-1000}\hbox{\strut attack}} & \vtop{\hbox{\strut Un-T }\hbox{\strut attack}} &\vtop{\hbox{\strut Multi-T}\hbox{\strut attack}}   & per batch \\
				\midrule
				\parbox[t]{2mm}{\multirow{4}{*}{\rotatebox[origin=c]{90}{CIFAR-10}}}
				& one-step ADV & 86.24$\%$ & 82.53$\%$ &45.73$\%$  & 44.98$\%$  & 45.14$\%$& 42.97$\%$  & 0.25s \\
                & one-step TRADES & $\boldsymbol{86.84}\%$ &$\boldsymbol{82.63}\%$ & 40.03$\%$ & 39.41$\%$ & 39.31$\%$  & 38.14$\%$& 0.55s \\
                & one-step ATLAS  & 84.52$\%$ &81.06$\%$ & $\boldsymbol{49.01}\%$& $\boldsymbol{48.59}\%$ & $\boldsymbol{48.54}\%$ & $\boldsymbol{46.55}\%$&  0.73s  \\
			    & one-step TradesJac  & 84.38$\%$ & 81.54$\%$ & 46.04$\%$& 45.75$\%$ & 44.60$\%$ & 43.59$\%$&  1.14s  \\
				\hline \hline  \TBstrut \TBstrut
				
				\parbox[t]{2mm}{\multirow{4}{*}{\rotatebox[origin=c]{90}{CIFAR-100}}}
				& one-step ADV & 48.88$\%$& 45.59$\%$  & 19.89$\%$ & 19.72$\%$ & 19.04$\%$ & - & 0.25s \\
                & one-step TRADES & $\boldsymbol{62.58}\%$ & $\boldsymbol{57.64}\%$ & 16.93$\%$ & 16.21$\%$ & 14.83$\%$  & -   & 0.53s \\
                & one-step ATLAS   & 60.98$\%$ & 57.04$\%$ & $\boldsymbol{26.30}\%$& $\boldsymbol{25.94}\%$ & $\boldsymbol{26.28}\%$ & -  & 0.73s  \\
			    & one-step TradesJac   & 58.15$\%$ & 54.56$\%$ & 21.22$\%$& 20.89$\%$ & 19.78$\%$ & -  & 1.15s  \\
			    \bottomrule
			\end{tabular}}
	\end{table*}

\section*{D.2 Additional attacks}
We perform two additional robustness evaluations on CIFAR-10 and CIFAR-100. The first one is a hard-label attack called RayS \citep{rays}. For both datasets, we run RayS with 40000 queries on 1000 randomly selected images from the test set. The second one is an ensemble of parameter-free attacks, named as AutoAttack \citep{AA}. AutoAttack consists of two variants of PGD attack, an attack focused on gradient-masking and a black-box attack. We have used the standard version of AutoAttack for all evaluations. Results for CIFAR-10 and CIFAR-100 can be found in Table \ref{table:AA-cifar10} and Table \ref{table:AA-cifar100} respectively. ATLAS gives the most robust performance among all methods in both weak and strong adversaries cases.
\begin{table}[h]
\parbox{.48\linewidth}{
\centering
	\rtable{1.2}
			\caption{\small{Robustness performance for CIFAR-10 on Wide-Resnet 28-8. The higher the better.}\label{table:AA-cifar10}}
\resizebox{0.9\linewidth}{!}{\begin{tabular}{clcc}
\toprule
                & Model & \vtop{\hbox{\strut RayS}\hbox{\strut attack }} &\vtop{\hbox{\strut Auto }\hbox{\strut attack}} \\
				\midrule
\parbox[t]{2mm}{\multirow{9}{*}{\rotatebox[origin=c]{90}{CIFAR-10}}}
				&  zero-step JAC & 33.50$\%$& 30.03$\%$ \\
				\cline{2-4}
				& one-step ADV & 50.20$\%$ & 42.49$\%$  \\
                & one-step TRADES & 45.10$\%$ & 37.74$\%$ \\
                & one-step ATLAS  & $\boldsymbol{52.70}\%$ & $\boldsymbol{46.16}\%$   \\
                & one-step TradesJac  & 50.20$\%$ & 43.16$\%$   \\
				\cline{2-4}
				 &ten-step ADV & $\boldsymbol{54.40}\%$ & 47.56$\%$ \\
                 &ten-step TRADES & 54.00$\%$ & 49.40$\%$ \\
                 &ten-step ATLAS & 53.20$\%$& $\boldsymbol{49.82}\%$ \\
                \bottomrule
\end{tabular}}
}
\hfill
\parbox{.48\linewidth}{
\centering
	\rtable{1.2}
			\caption{\small{Robustness performance for CIFAR-100 on Wide-Resnet 28-8. The higher the better.} \label{table:AA-cifar100}}
\resizebox{0.9\linewidth}{!}{\begin{tabular}{clcc}
\toprule
                & Model & \vtop{\hbox{\strut RayS}\hbox{\strut attack }} &\vtop{\hbox{\strut Auto }\hbox{\strut attack}} \\
				\midrule
\parbox[t]{2mm}{\multirow{8}{*}{\rotatebox[origin=c]{90}{CIFAR-100}}}
				&  zero-step JAC & 16.70$\%$& 11.85$\%$ \\
				\cline{2-4}
				& one-step ADV & 19.90$\%$ & 17.17$\%$  \\
                & one-step TRADES & 19.50$\%$ & 12.75$\%$ \\
                & one-step ATLAS  & $\boldsymbol{28.60}\%$ & $\boldsymbol{23.42}\%$   \\
                & one-step TradesJac  & 22.60$\%$ & 17.99$\%$   \\
				\cline{2-4}
				 &ten-step ADV & 29.80$\%$ & 24.64$\%$ \\
                 &ten-step TRADES & 28.20$\%$ & 22.78$\%$ \\
                 &ten-step ATLAS & $\boldsymbol{30.60}\%$& $\boldsymbol{25.77}\%$ \\
                \bottomrule
\end{tabular}}
}
\end{table}

\section*{E Catastrophic Overfitting}
\begin{figure}[h!]
\centering
 \subfloat[]{\includegraphics[width=0.3\textwidth]{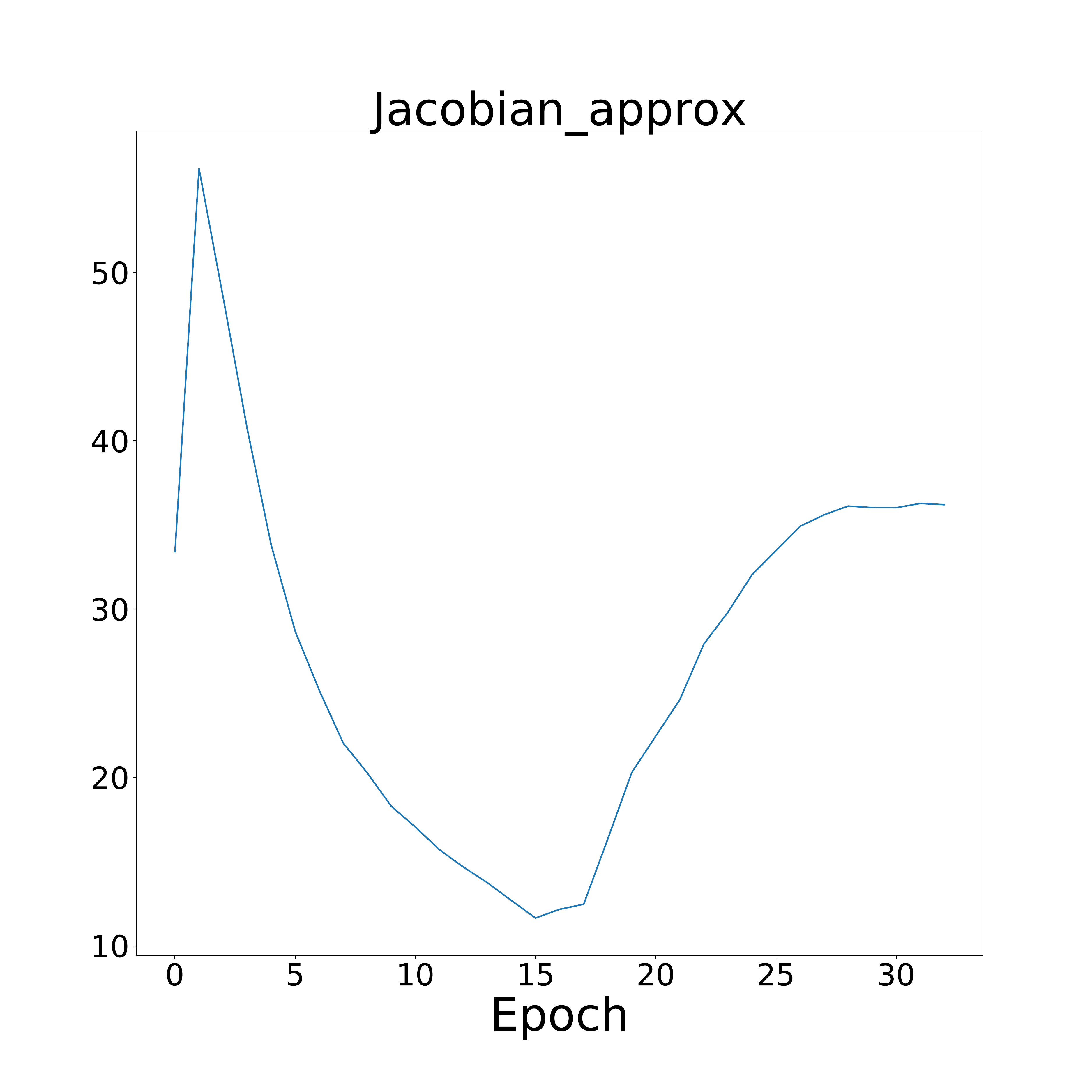}} 
    \subfloat[]{\includegraphics[width=0.3\textwidth]{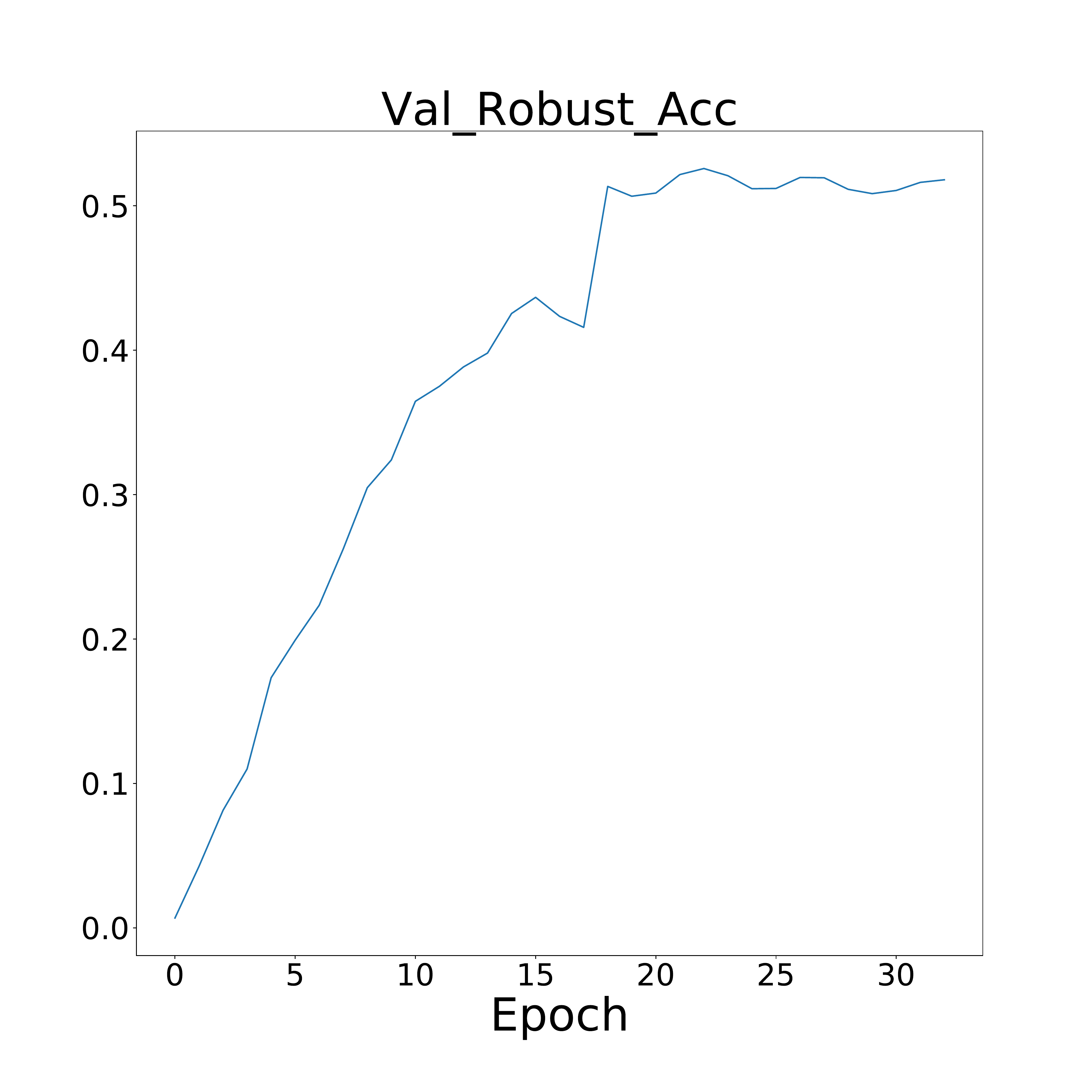}} 
    \subfloat[]{\includegraphics[width=0.3\textwidth]{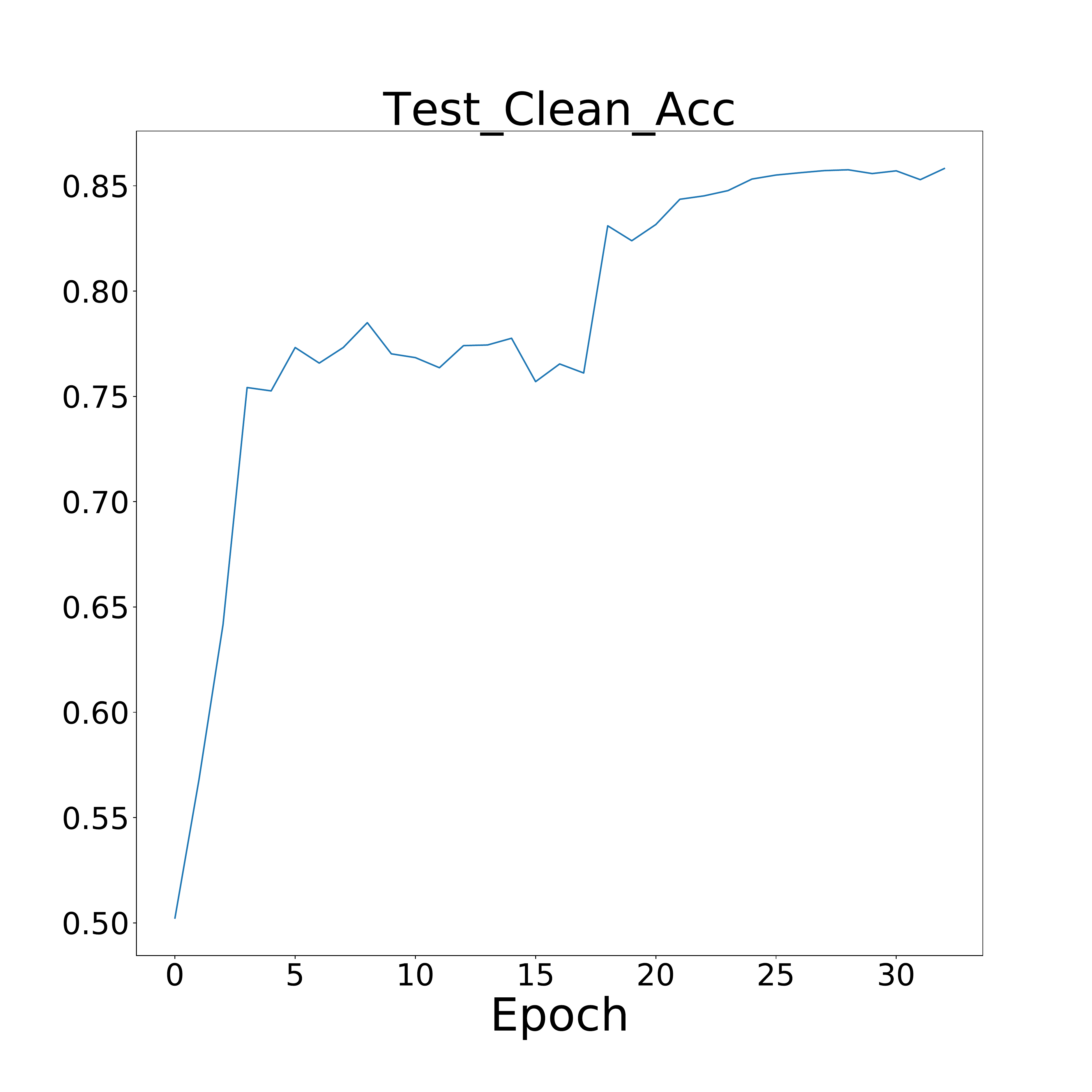}}
  \LineSep
 \subfloat[]{\includegraphics[width=0.3\textwidth]{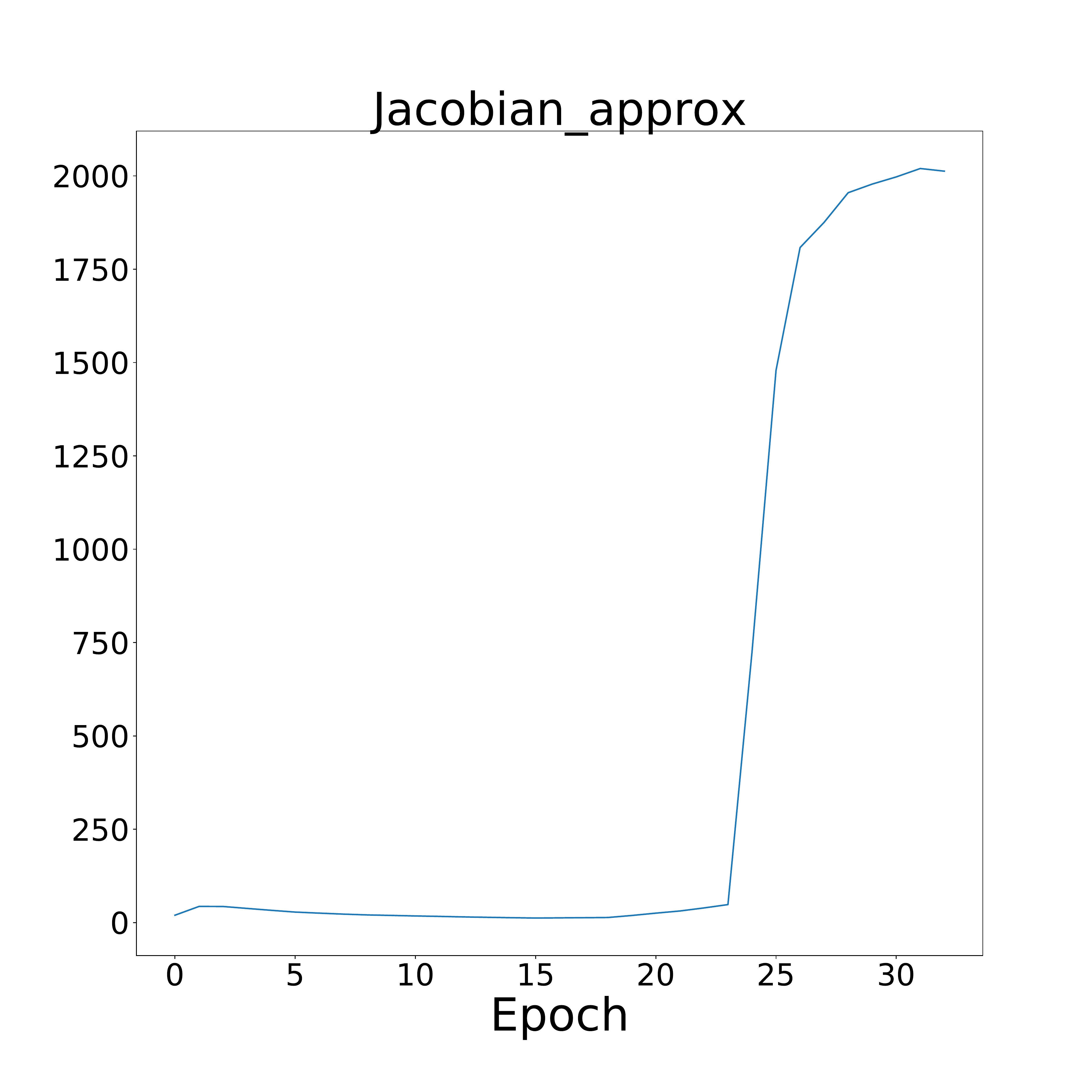}} 
    \subfloat[]{\includegraphics[width=0.3\textwidth]{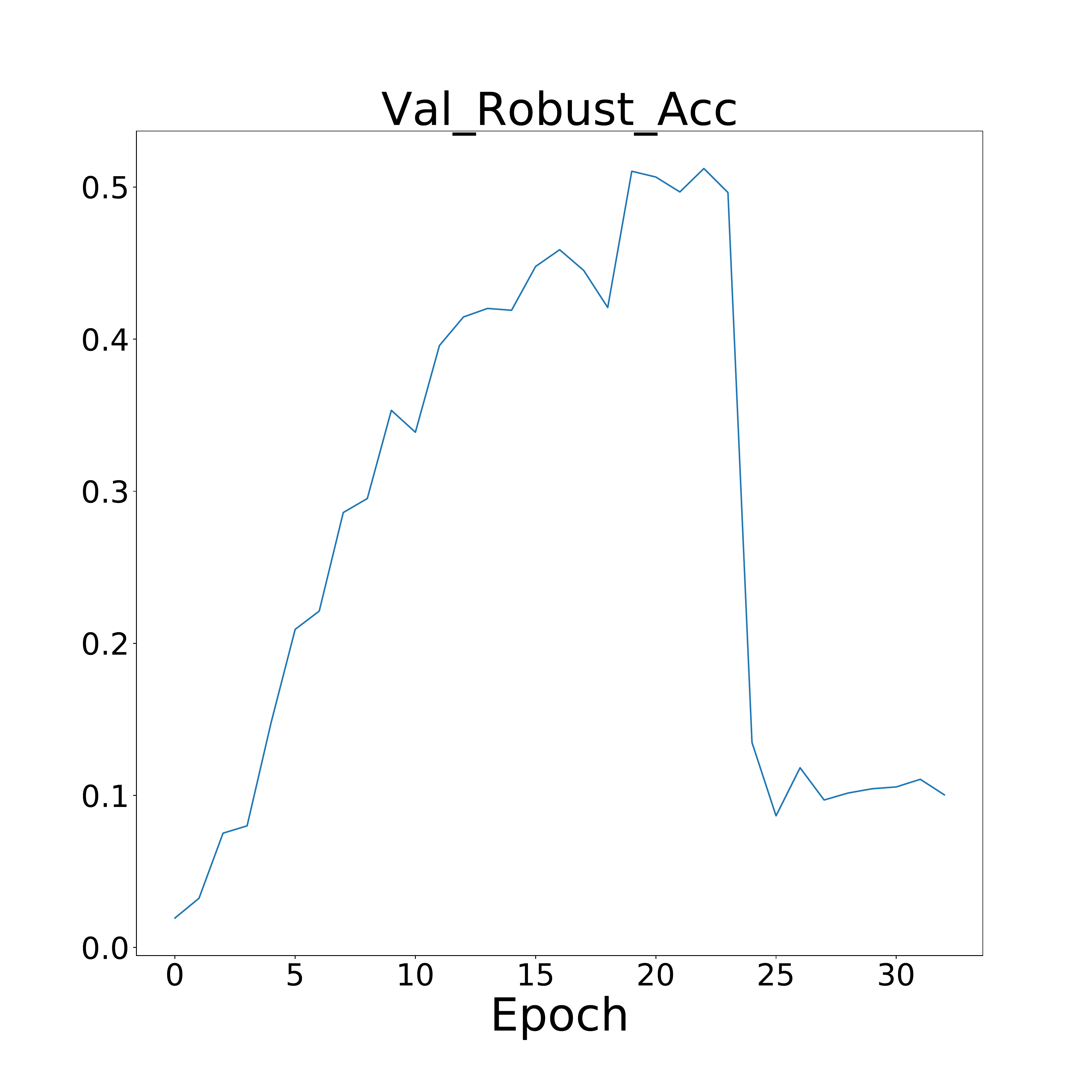}} 
    \subfloat[]{\includegraphics[width=0.3\textwidth]{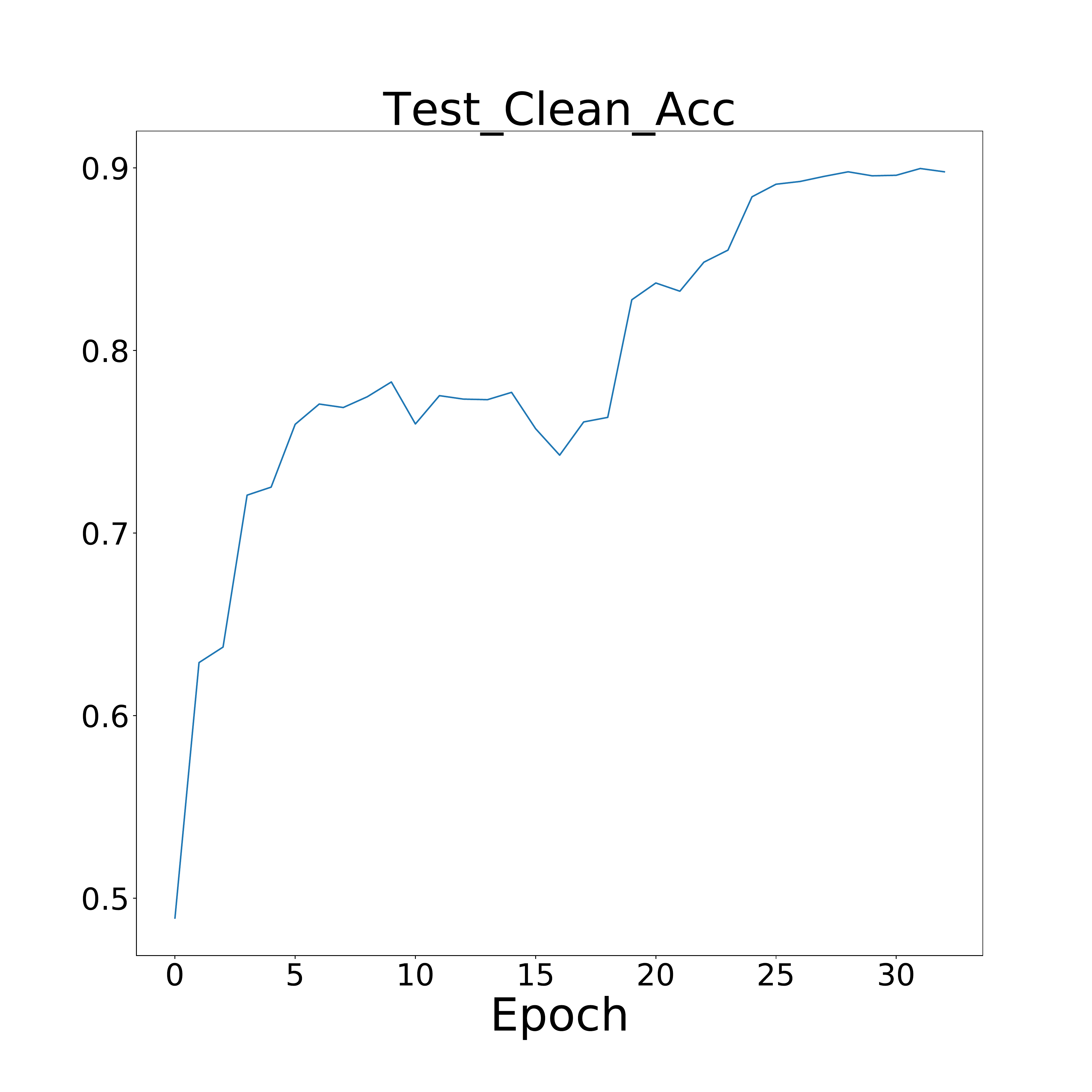}}
    \caption{ We show typical training plots for a model on CIFAR-10 when catastrophic overfitting is avoided in the upper row and when catastrophic overfitting happens in the bottom row. Upper row plots are for the ATLAS model with $\alpha=0.0002$ and $\beta=5.0$ while the bottom row plots are for the ATLAS model with $\alpha=0$ and $\beta=8.$. The figures on the left shows the value of $\Vert J(\boldsymbol{x}')\Vert^{\text{approx}}_F$ throughout the training. Catastrophic overfitting is associated with a steep increase of the Jacobian value. The figures in the middle gives the trend for robust accuracy on the validation dataset. Validation robust accuracy drops sharply with the sudden increase of the Jacobian value. Finally, on the right, we have plots for clean accuracy on the test set after each epoch. A further increase to almost $90\%$ clean accuracy is experienced in the catastrophic overfitting case.}
    \label{fig:co-bad}
\end{figure}

We give more details on catastrophic overfitting behaviour in our experiments. We first show typical training plots for both cases. The upper row of Figure \ref{fig:co-bad} contains plots for a model when catastrophic overfitting is avoided and the bottom row show plots for a model when catastrophic overfitting happens. It is clear to see that catastrophic overfitting is highly associated with a sharp increase of Jacobian value, which is consistent with the observations made in \citep{understanding2020}. In our experiments with roughly 30 epochs, we find that catastrophic overfitting always occurs for one-step TRADES and ATLAS-g but less frequently for one-step ADV and one-step ATLAS. For ATLAS-l, the catastrophic overfitting can be effectively avoided by increasing the value for $\alpha$ as it directly penalizes large Jacobian value. The same strategy can be applied for the full ATLAS method. We find that, for a stable training performance, the choice of $\alpha$ depends on the value of $\beta$. That is a large $\beta$ value requires a large $\alpha$ value. 

Finally, we mention that since the goal is to achieve high robust accuracy, we use early termination and consider a wider range of models regardless whether they experience catastrophic overfitting in the end.   

\section*{F Parameter Choices: One-Step Case}

We show clean accuracy and robust accuracy for each method at various parameter choices. 

\subsection*{F.1 Zero-Step Jac}
\begin{figure}[h]
  \centering
   \begin{minipage}[b]{0.45\textwidth}
    \includegraphics[width=0.6\textwidth]{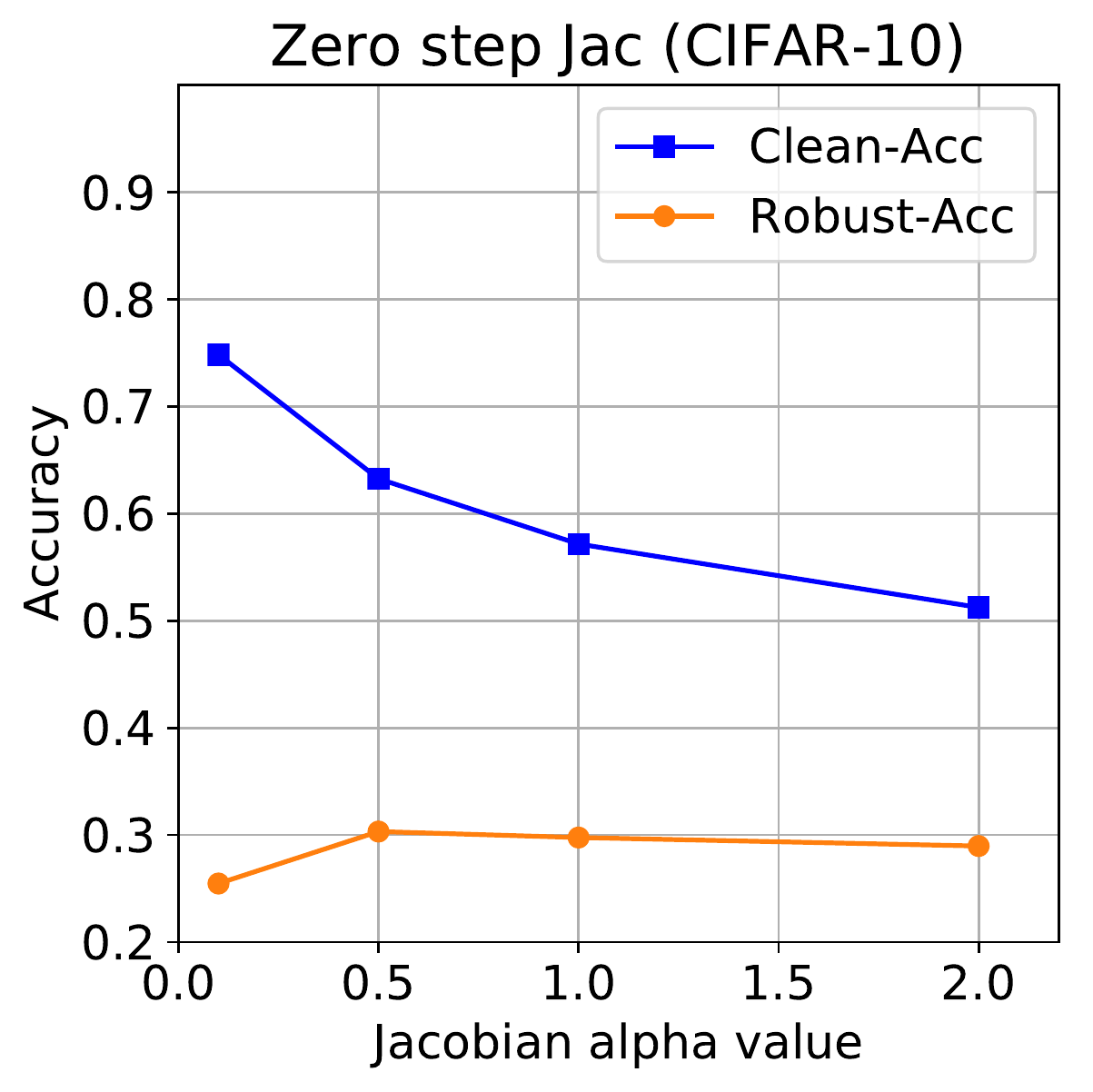}
    \caption{Clean and robust accuracy for zero-step JAC at different $\alpha_{\text{Jac}}$ on CIFAR-10}
    \label{fig:zerostepjacc10}
  \end{minipage}
  \hfill
  \begin{minipage}[b]{0.45\textwidth}
    \includegraphics[width=0.6\textwidth]{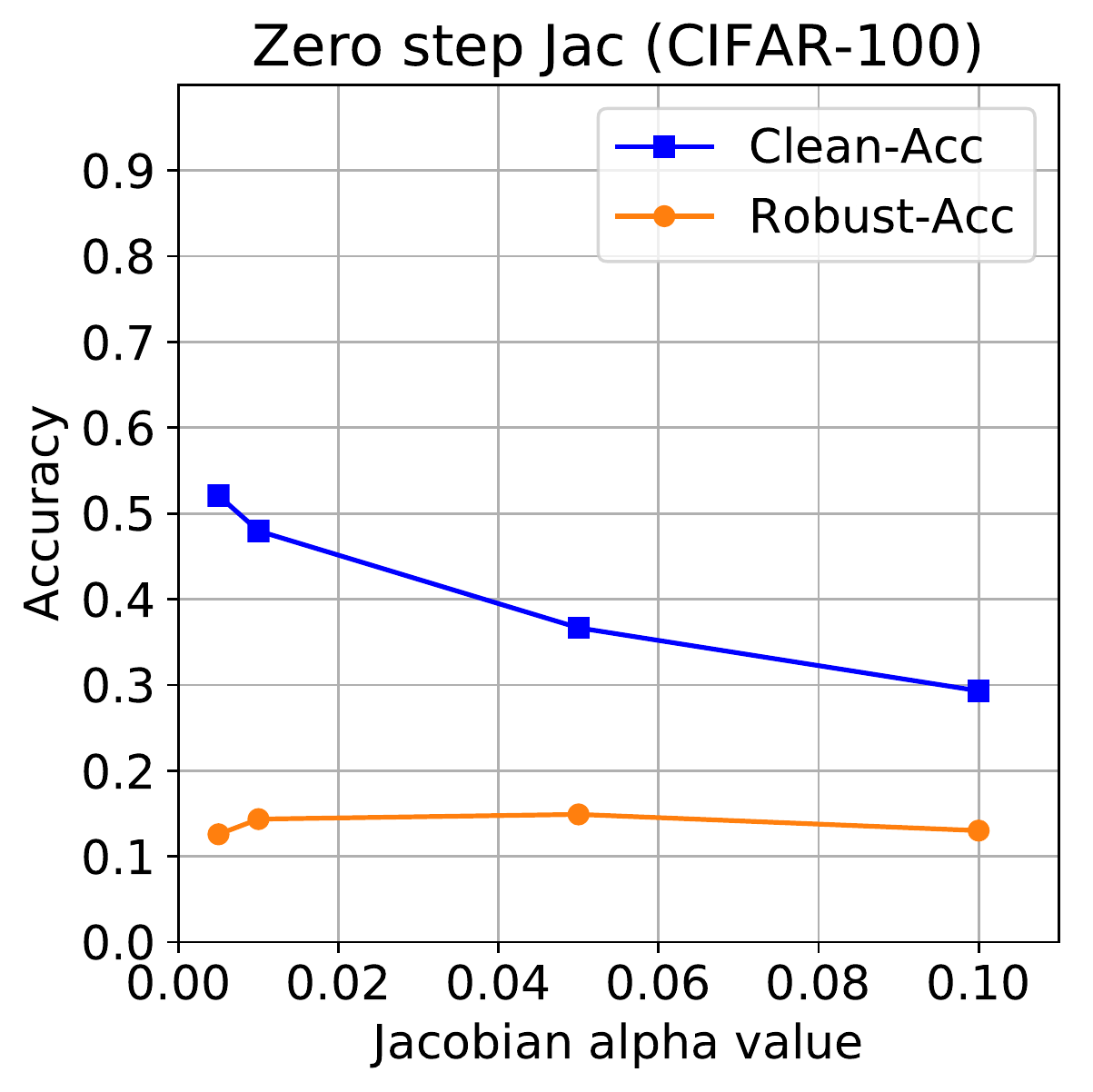}
    \caption{Clean and robust accuracy for zero-step JAC at different $\alpha_{\text{Jac}}$ on CIFAR-100}
    \label{fig:zerostepjacc100}
  \end{minipage}
\end{figure}

Although zero-step Jac does not require adversaries, we used 2 epochs for adjusting learning rate via validation robust accuracy to be consistent. Model's performance over a range of $\alpha_{\text{Jac}}$ values is shown in Figure~\ref{fig:zerostepjacc10} for CIFAR-10 and \ref{fig:zerostepjacc100} for CIFAR-100. There is a large trade-off between clean and robust accuracy for small $\alpha_{\text{Jac}}$ values and then both clean and robust accuracy decrease with the increase of $\alpha_{\text{Jac}}$.

\subsection*{F.2 One-Step TRADES}

\begin{figure}[h]
  \centering
   \begin{minipage}[b]{0.45\textwidth}
    \includegraphics[width=0.6\textwidth]{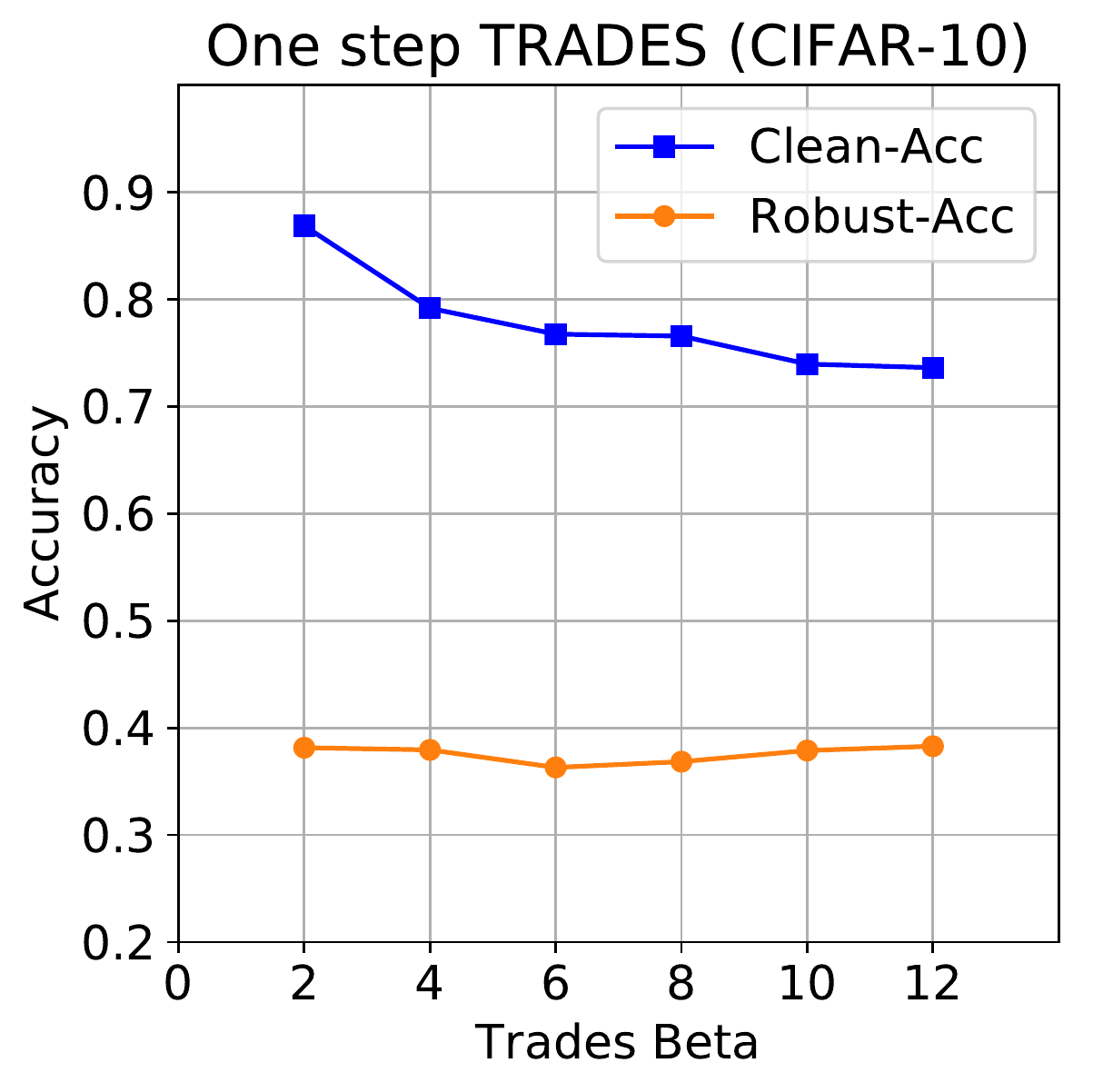}
    \caption{Clean and robust accuracy for one-step TRADES at different $\beta_{\text{TRADES}}$ on CIFAR-10}
    \label{fig:onesteptradesc10}
  \end{minipage}
  \hfill
  \begin{minipage}[b]{0.45\textwidth}
    \includegraphics[width=0.6\textwidth]{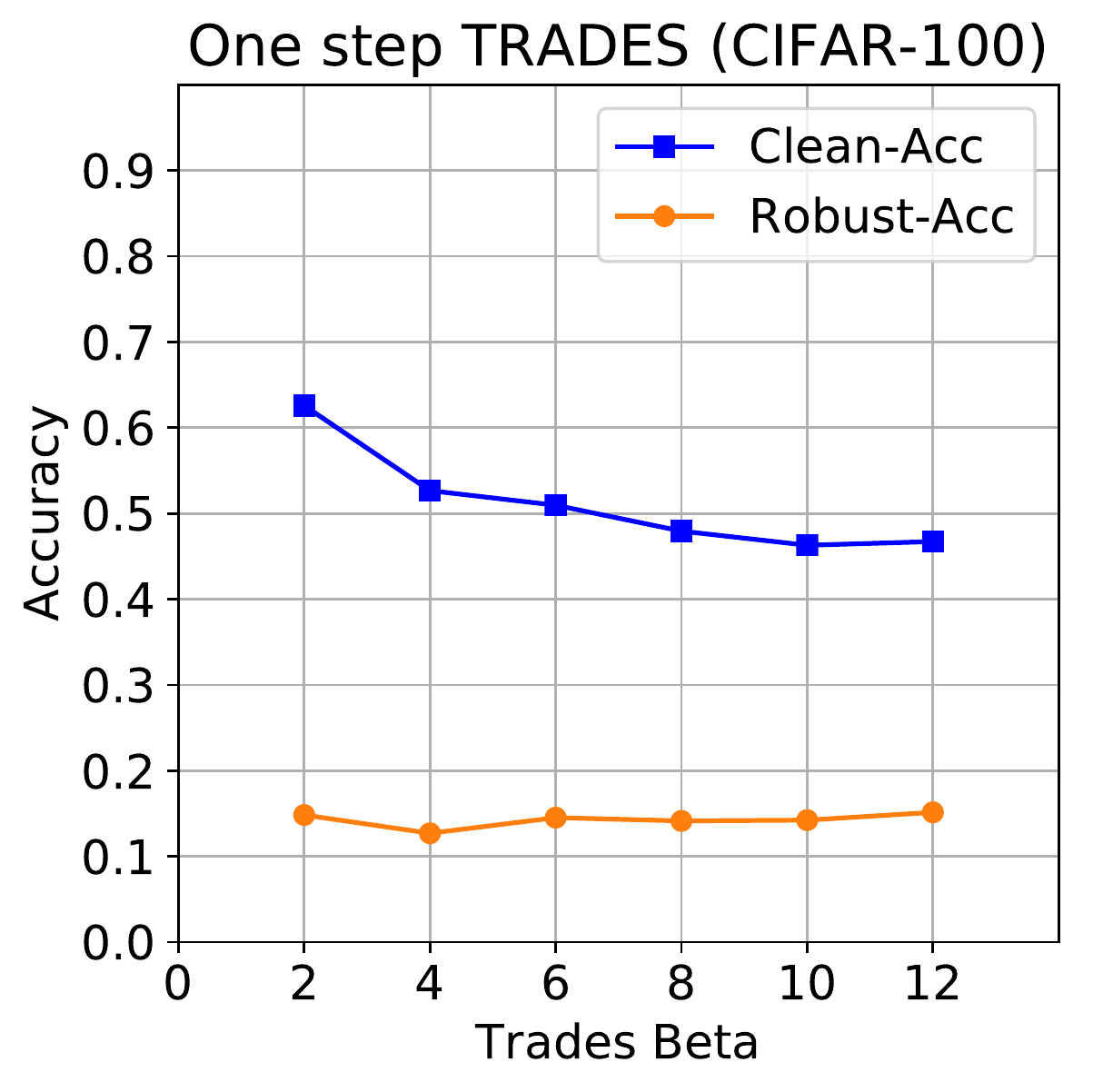}
    \caption{Clean and robust accuracy for one-step TRADES at different $\beta_{\text{TRADES}}$ on CIFAR-100}
    \label{fig:onesteptradesc100}
  \end{minipage}
\end{figure}

In Figure~\ref{fig:onesteptradesc10} (CIFAR-10) and \ref{fig:onesteptradesc100} (CIFAR-100), we show one-step TRADES at various $\beta_{\text{TRADES}}$ values. It is easy to see that increasing the value of $\beta_{\text{TRADES}}$ mainly hurts the clean accuracy without improving robust accuracy. 

\subsection*{F.3 One-Step ATLAS}
\begin{figure}[h]
\vspace{-10pt}
\centering
 \subfloat[]{\includegraphics[width=0.25\textwidth]{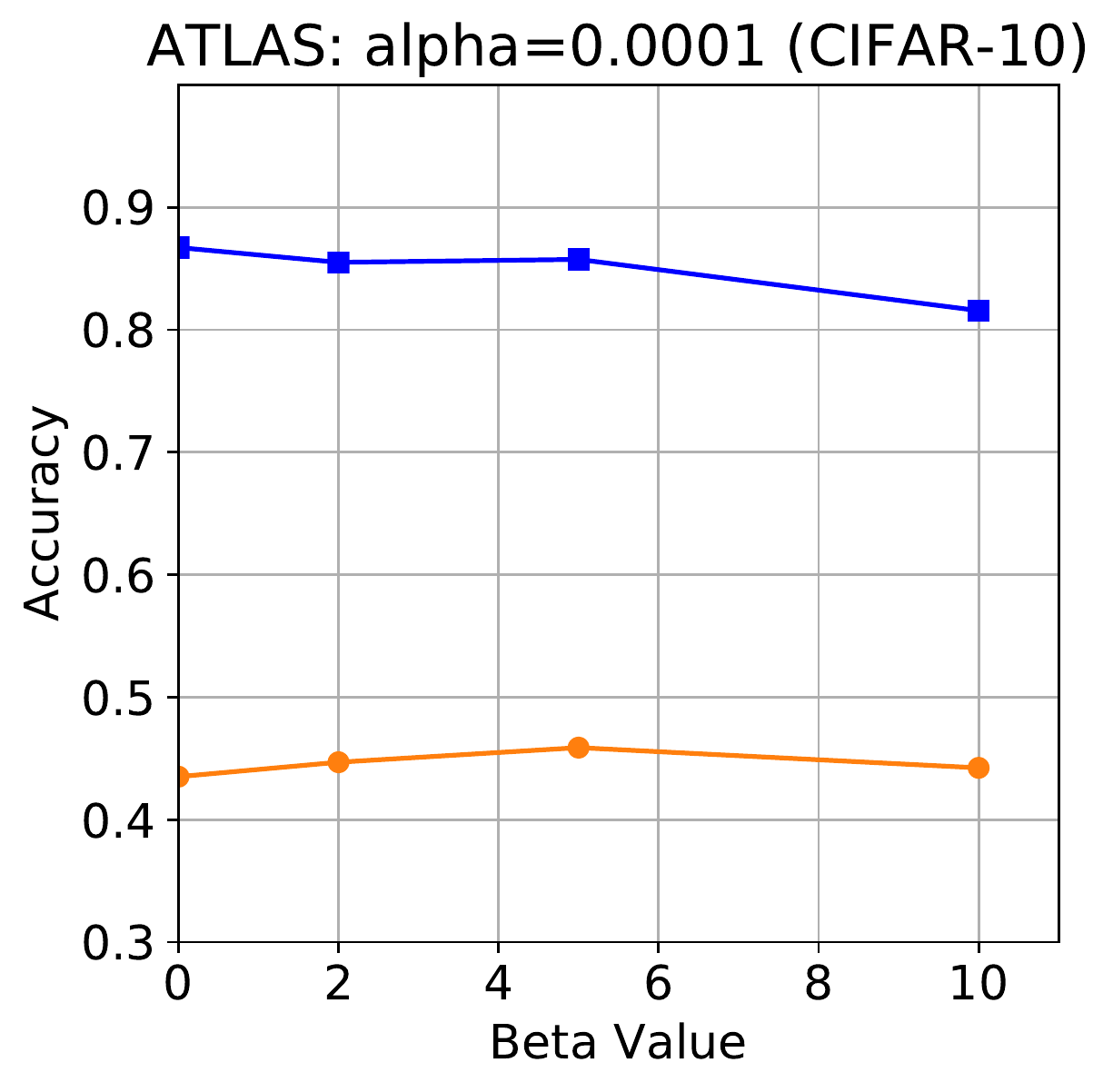}} 
    \subfloat[]{\includegraphics[width=0.25\textwidth]{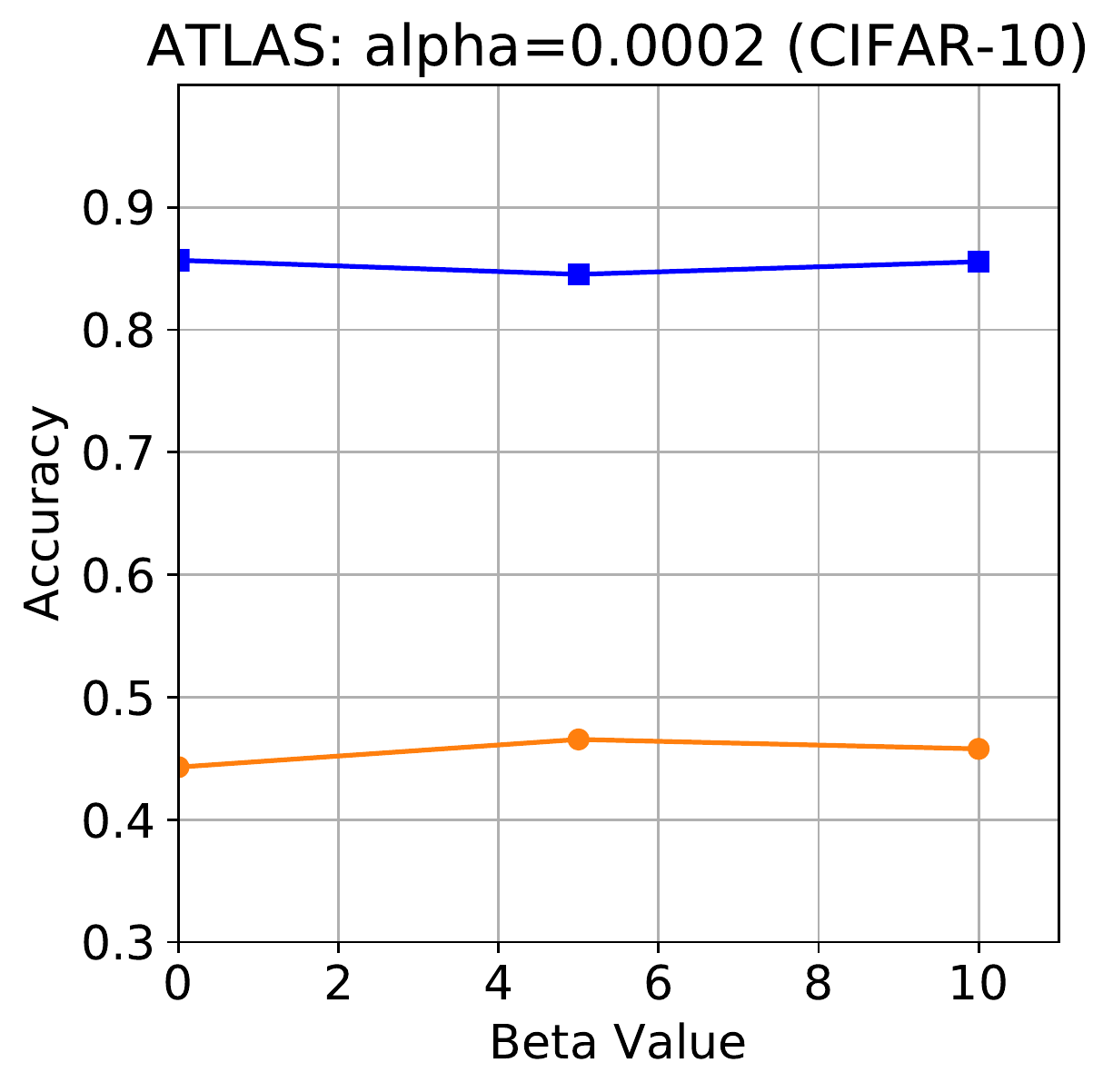}} 
    \subfloat[]{\includegraphics[width=0.36\textwidth]{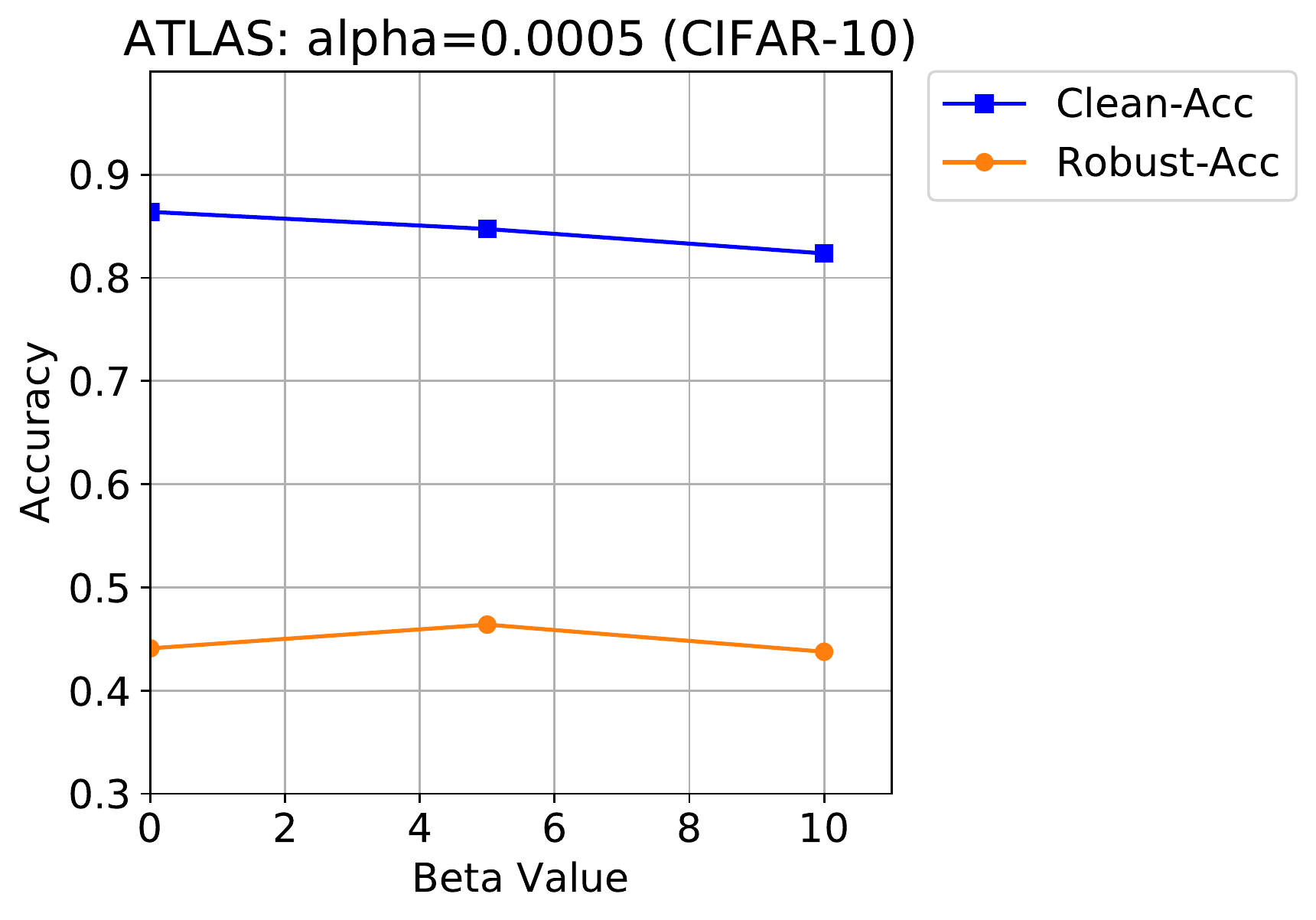}}
\LineSep
 \subfloat[]{\includegraphics[width=0.25\textwidth]{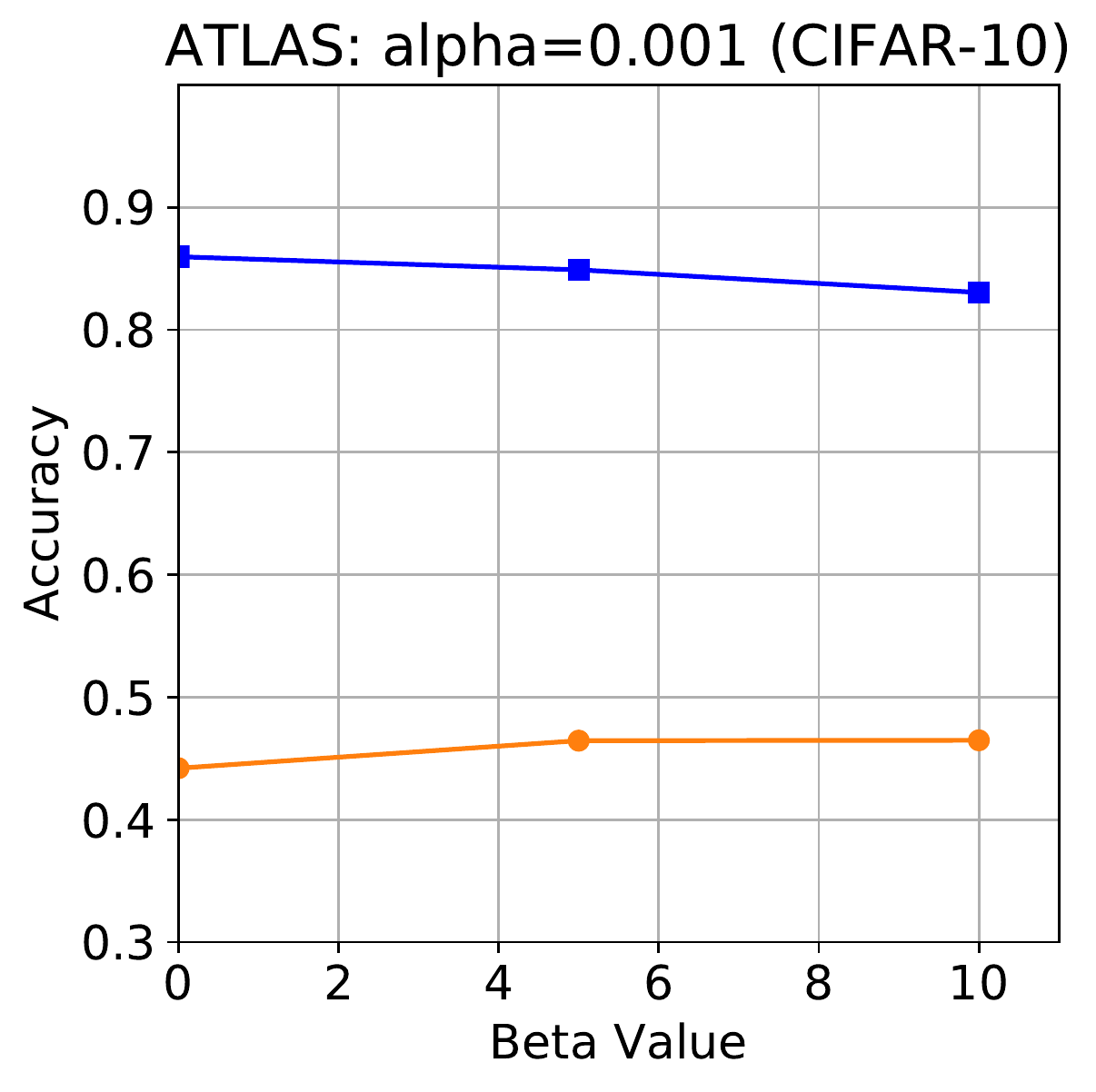}} 
    \subfloat[]{\includegraphics[width=0.25\textwidth]{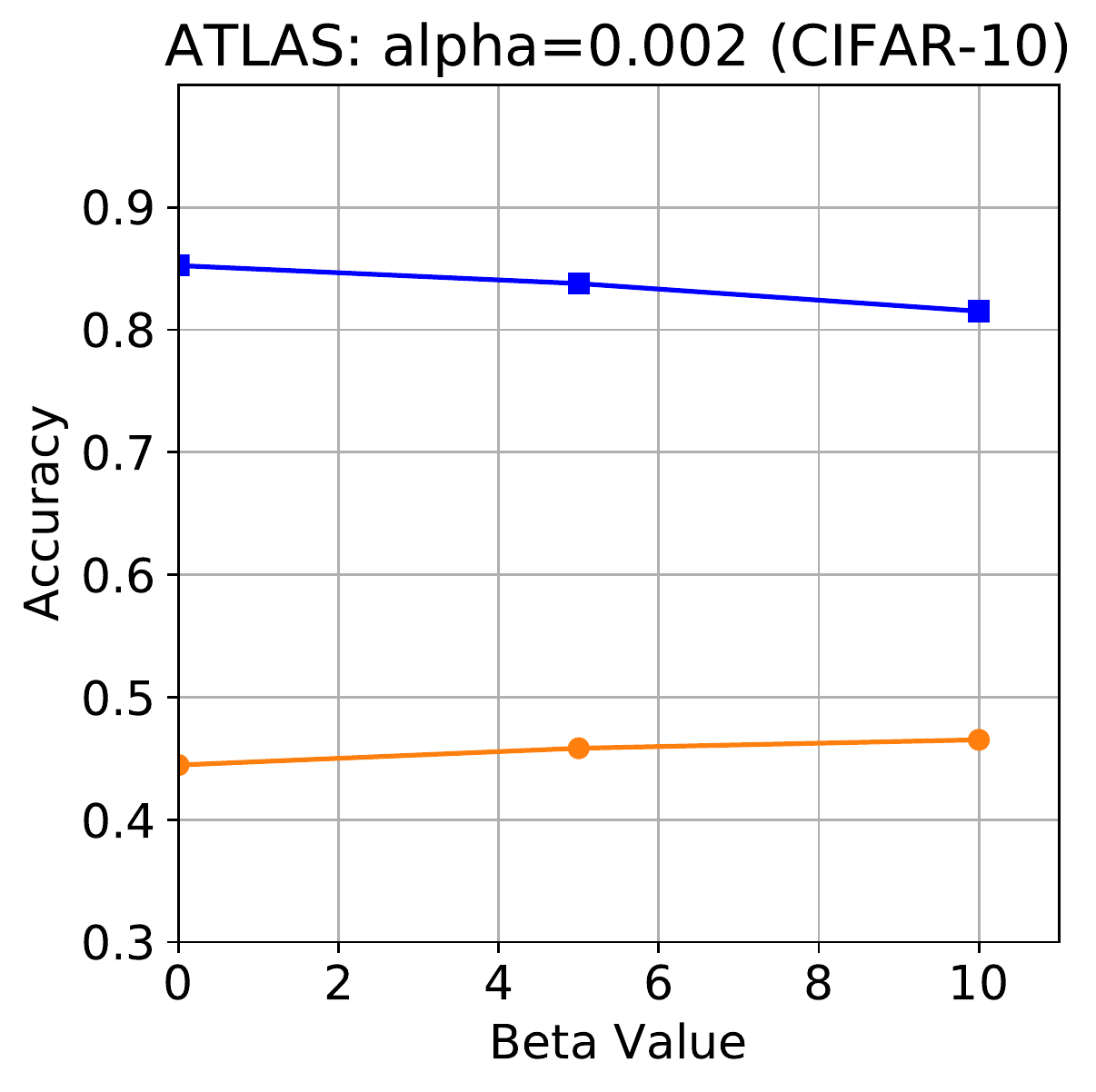}} 
    \subfloat[]{\includegraphics[width=0.36\textwidth]{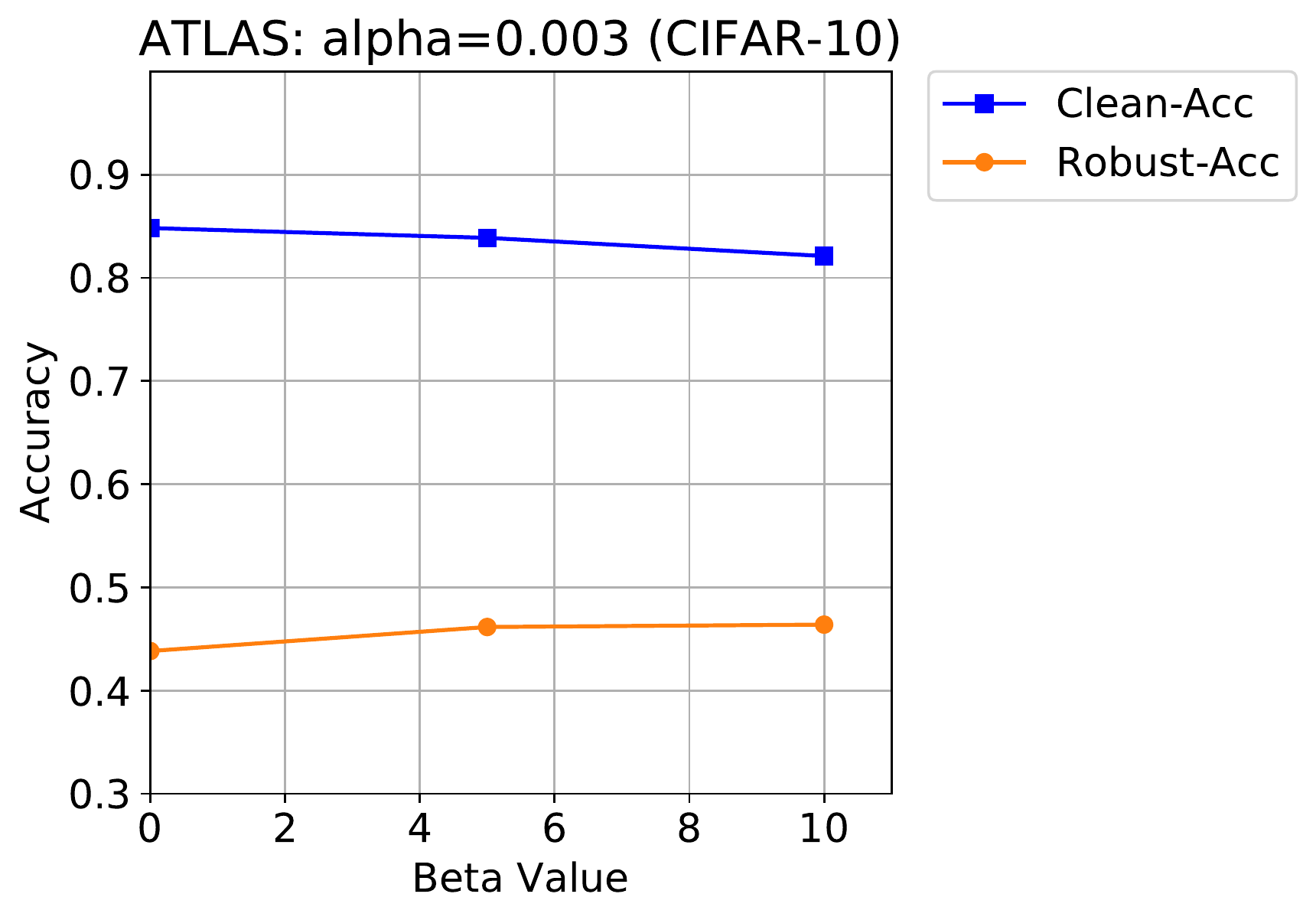}}
    \caption{Clean and robust accuracy for one-step ATLAS at different $\alpha, \beta$ on CIFAR-10}
    \label{fig:one-NRL-c10}
\end{figure}

\begin{figure}[htbp]

\centering
 \subfloat[]{\includegraphics[width=0.24\textwidth]{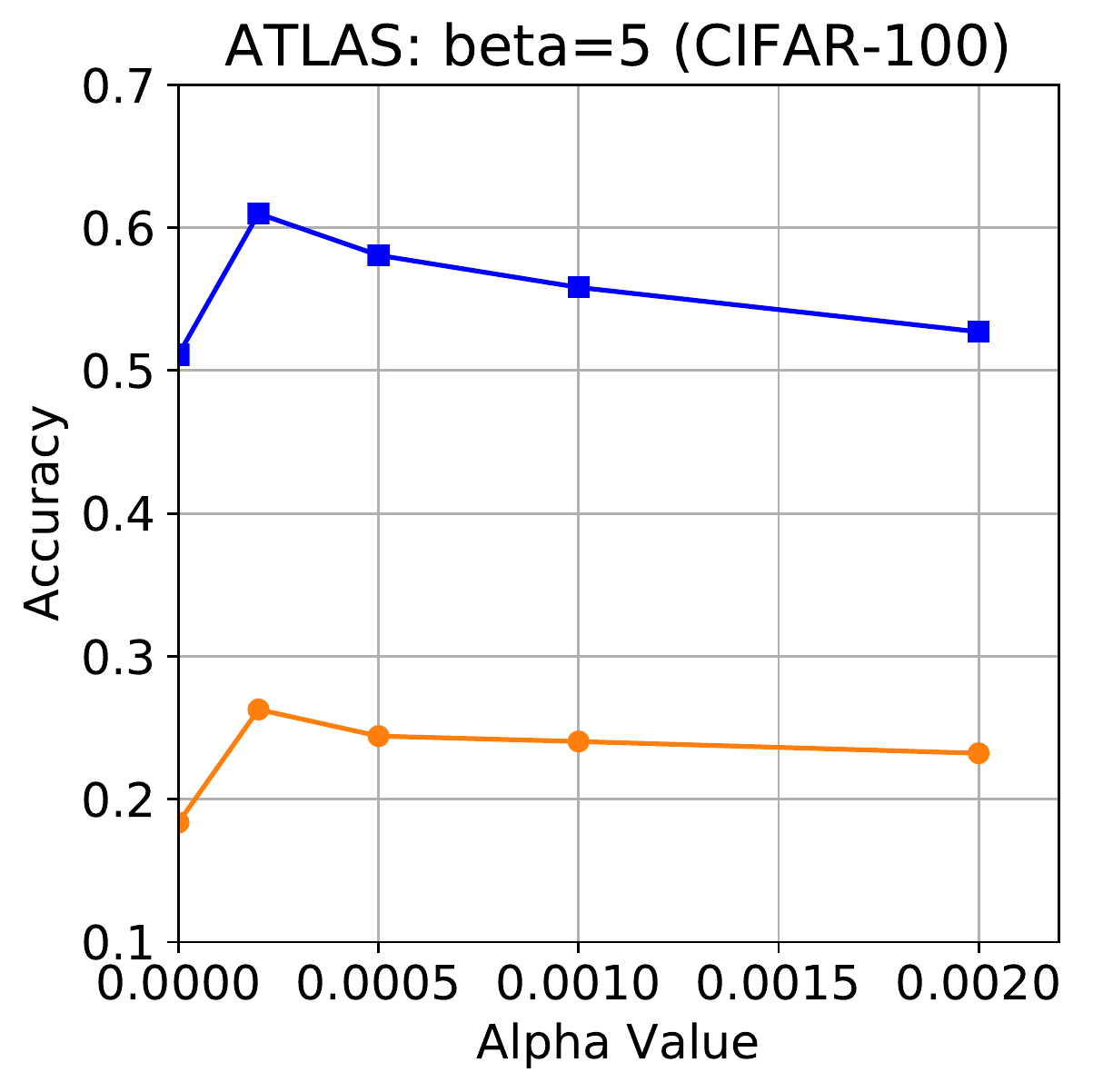}} 
    \subfloat[]{\includegraphics[width=0.24\textwidth]{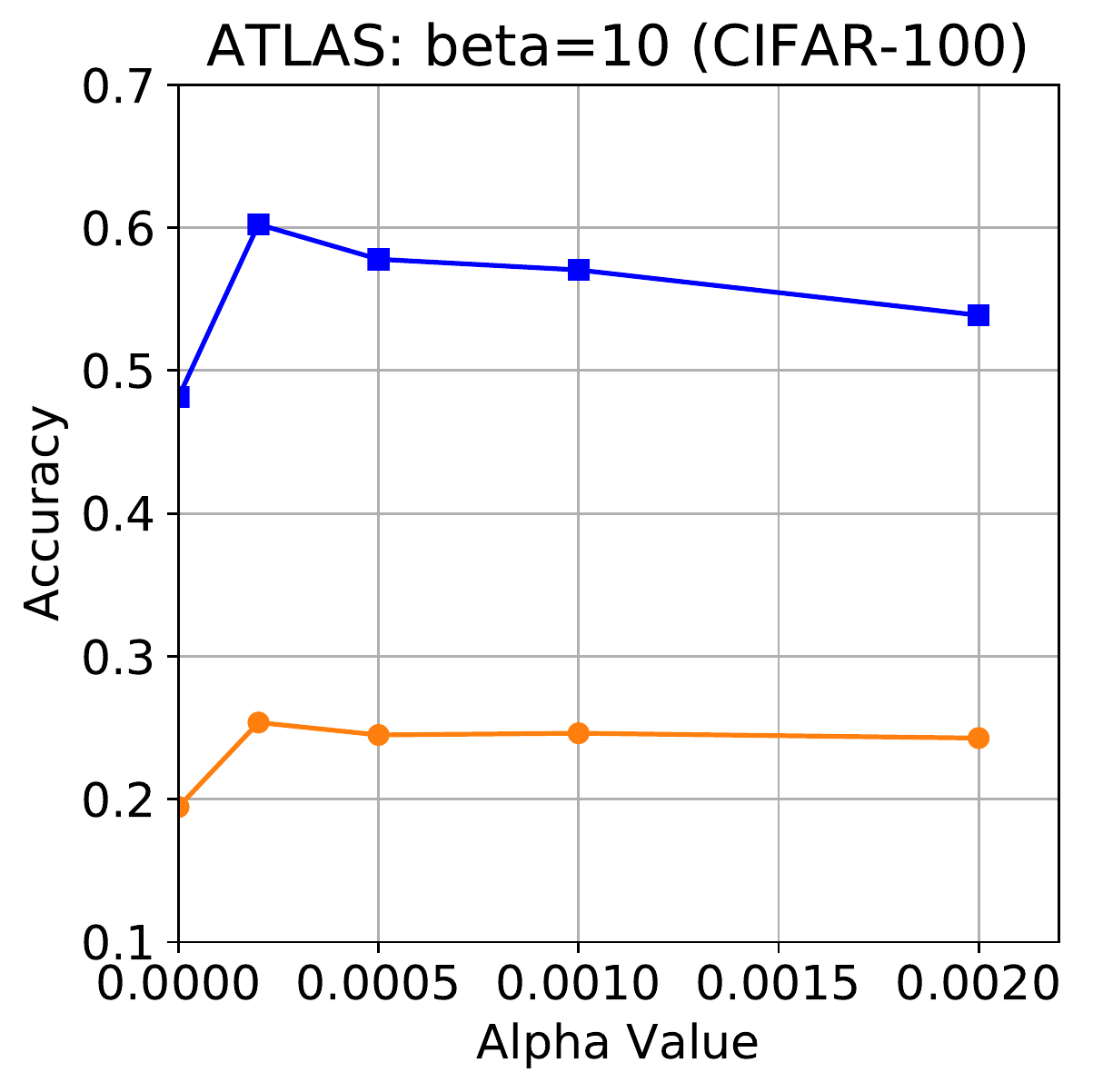}} 
    \subfloat[]{\includegraphics[width=0.24\textwidth]{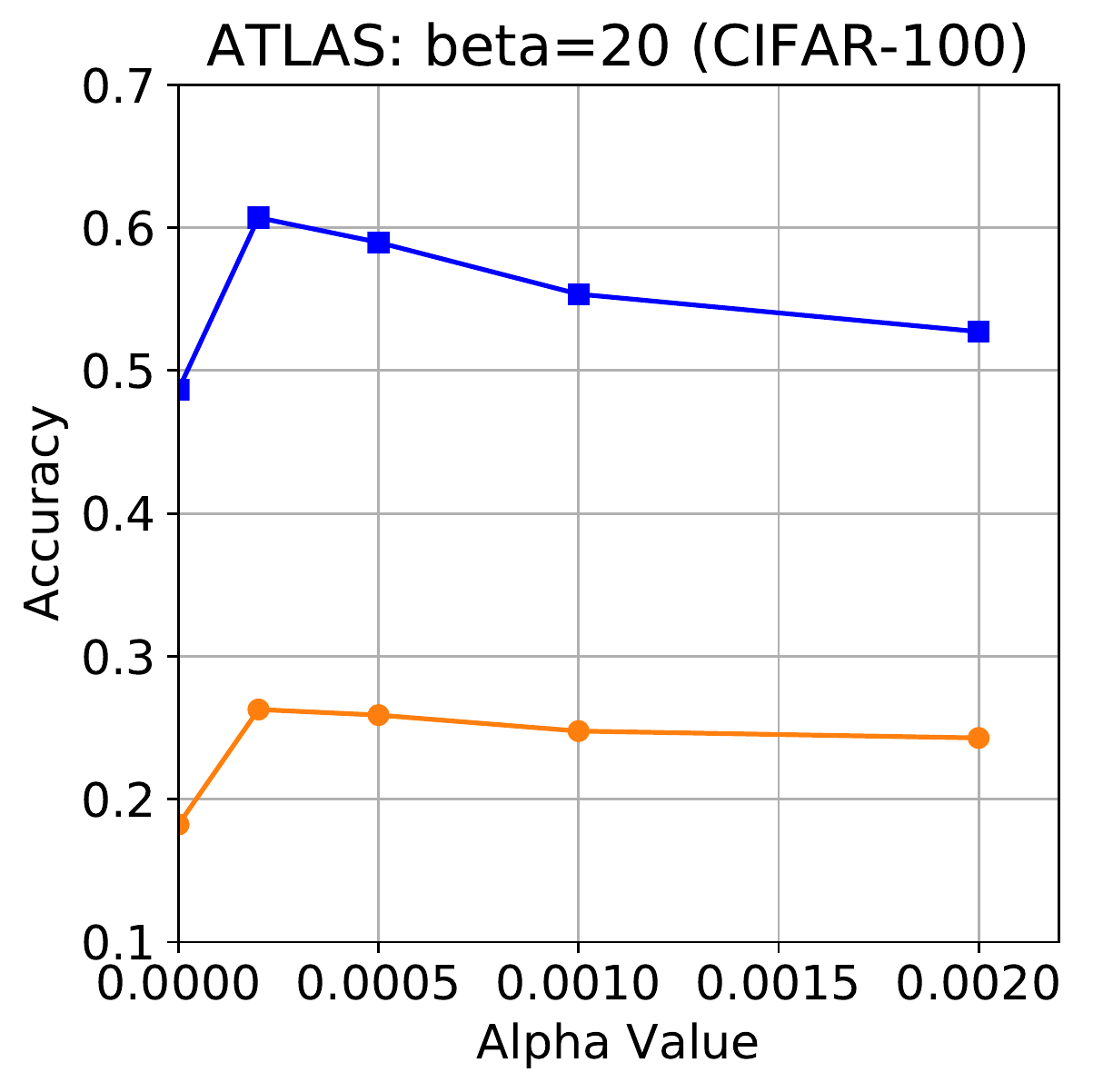}}
    \subfloat[]{\includegraphics[width=0.25\textwidth]{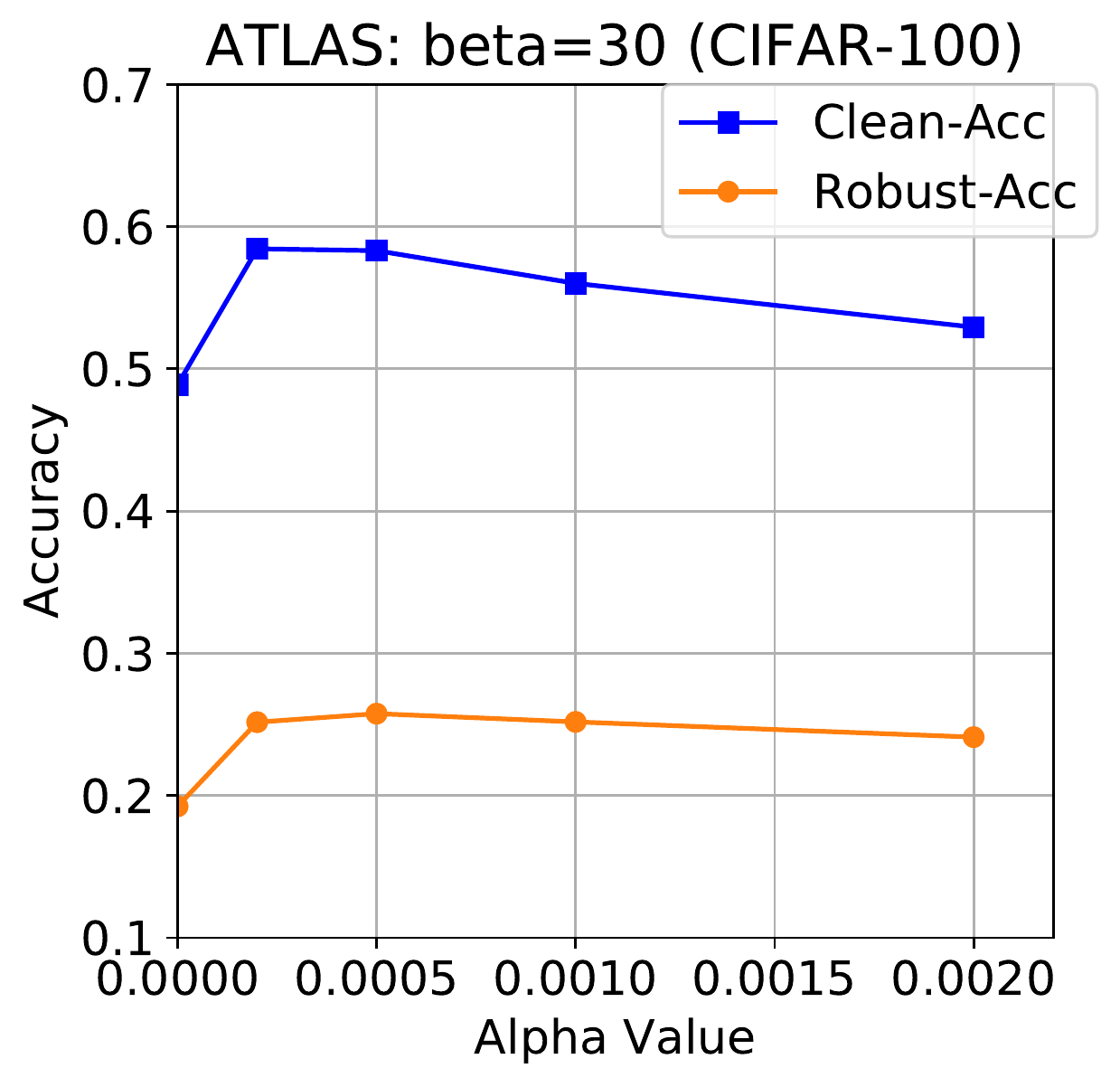}}
    \caption{Clean and robust accuracy for one-step ATLAS at different $\alpha, \beta$ on CIFAR-100}
    \label{fig:one-NRL-c100}
\end{figure}

In Figure~\ref{fig:one-NRL-c10}, we give clean and robust accuracy for models trained at different $\alpha$ and $\beta$ values on CIFAR-10. In Figure~\ref{fig:one-NRL-c100}, we give clean and robust accuracy for models trained at different $\alpha$ and $\beta$ values on CIFAR-100. A slight trade-off between clean and robust accuracy can be observed. 

\end{document}